\documentclass[10pt,journal,compsoc]{IEEEtran}

\usepackage{url}

\usepackage{amsfonts}

\usepackage{amssymb}

\ifCLASSOPTIONcompsoc
  \usepackage[nocompress]{cite}
\else
  \usepackage{cite}
\fi
\ifCLASSINFOpdf
\else
\fi

\usepackage{enumerate}
\usepackage{algorithm}
\usepackage{multirow}
\usepackage{algpseudocode}
\usepackage{lineno}
\usepackage{xcolor}
\usepackage{subfigure}
\usepackage{caption}
\usepackage{multirow}
\usepackage{float}
\usepackage{cite}
\usepackage{graphicx}
\usepackage{amsmath}
\usepackage{array}
\usepackage{threeparttable}
\ifCLASSOPTIONcompsoc
 \usepackage[caption=false,font=normalsize,labelfont=sf,textfont=sf]{subfig}
\else
 \usepackage[caption=false,font=footnotesize]{subfig}
\fi

\usepackage{fixltx2e}

\usepackage{dblfloatfix}

\ifCLASSOPTIONcaptionsoff
  \usepackage[nomarkers]{endfloat}
 \let\MYoriglatexcaption\caption
 \renewcommand{\caption}[2][\relax]{\MYoriglatexcaption[#2]{#2}}
\fi

\usepackage{url}
\usepackage{ragged2e}

\begin{document}
%
\title{Texture variation adaptive image denoising with nonlocal PCA}
%
%
%

\author{Wenzhao Zhao, Qiegen Liu, Yisong Lv, and Binjie Qin*,~\IEEEmembership{Member,~IEEE}
\thanks{Manuscript received April 14, 2017; accepted October 27, 2017; This work was supported by the National Natural Science Foundation of China (61271320 and 61503176), the young scientist training plan of Jiangxi province under grant 20162BCB23019, and Medical Engineering Cross Fund of Shanghai Jiao Tong University (YG2014MS29).}
\thanks{Wenzhao Zhao and Binjie Qin are with the School of Biomedical Engineering, Shanghai Jiao Tong University, Shanghai, 200240, China. Qiegen Liu is with the Department of Electronic Information Engineering, Nanchang University, Nanchang 330031, China. Yisong Lv is with School of Mathematical Sciences, Shanghai Jiao Tong University, Shanghai 200240, China. (e-mail: bjqin@sjtu.edu.cn).}}

%
%

\markboth{Journal of \LaTeX\ Class Files,~Vol.~14, No.~8, August~2017}%
{Shell \MakeLowercase{\textit{et al.}}: Bare Advanced Demo of IEEEtran.cls for IEEE Computer Society Journals}
\IEEEtitleabstractindextext{%
\begin{abstract}
\justifying
Image textures, as a kind of local variations, provide important information for human visual system. Many image textures, especially the small-scale or stochastic textures are rich in high-frequency variations, and are difficult to be preserved. Current state-of-the-art denoising algorithms typically adopt a nonlocal approach consisting of image patch grouping and group-wise denoising filtering. To achieve a better image denoising while preserving the variations in texture, we first adaptively group high correlated image patches with the same kinds of texture elements (texels) via an adaptive clustering method. This adaptive clustering method is applied in an over-clustering-and-iterative-merging approach, where its noise robustness is improved with a custom merging threshold relating to the noise level and cluster size. For texture-preserving denoising of each cluster, considering that the variations in texture are captured and wrapped in not only the between-dimension energy variations but also the within-dimension variations of PCA transform coefficients, we further propose a PCA-transform-domain variation adaptive filtering method to preserve the local variations in textures. Experiments on natural images show the superiority of the proposed transform-domain variation adaptive filtering to traditional PCA-based hard or soft threshold filtering. As a whole, the proposed denoising method achieves a favorable texture preserving performance both quantitatively and visually, especially for stochastic textures, which is further verified in camera raw image denoising.
\justifying
\end{abstract}


}

\maketitle

\IEEEdisplaynontitleabstractindextext

%
\IEEEpeerreviewmaketitle

\ifCLASSOPTIONcompsoc
\IEEEraisesectionheading{\section{Introduction}\label{sec:introduction}}
\else
\section{Introduction}
\label{sec:introduction}
\fi

%
%
%
%
\IEEEPARstart{T}exture, as a systematic local variation of image values, is an essential component of visual information reflecting the physical properties of the surrounding environment \cite{haidekker2011texture}. There are two basic types of texture pattern: regular texture that consisting of repeated texture elements (texels) and stochastic texture without explicit texels\cite{Texturesynthesis}. Most of the real-world textures locate in-between these two extremes.

Preservation of texture variation is necessary for image preprocessing tasks such as image denoising, so as to help make better use of the image content. However, texture variation, especially the small-scale or stochastic texture variation often lies in high frequency bands. These high frequency variations are  difficult to be preserved during noise removal and tends to be smoothed. The existing state-of-the-art denoising methods often adopt the nonlocal methodology, which firstly uses patch grouping (PG) techniques to exploit the nonlocal self-similarity (NSS) prior in natural images, and then uses denoising filters (DF) for group-wise denoising. Over-smoothness of the image textures is caused by the deficiencies in both PG and DF procedures.

\begin{figure}[htb]
 \center{\includegraphics[width=8.1cm]  {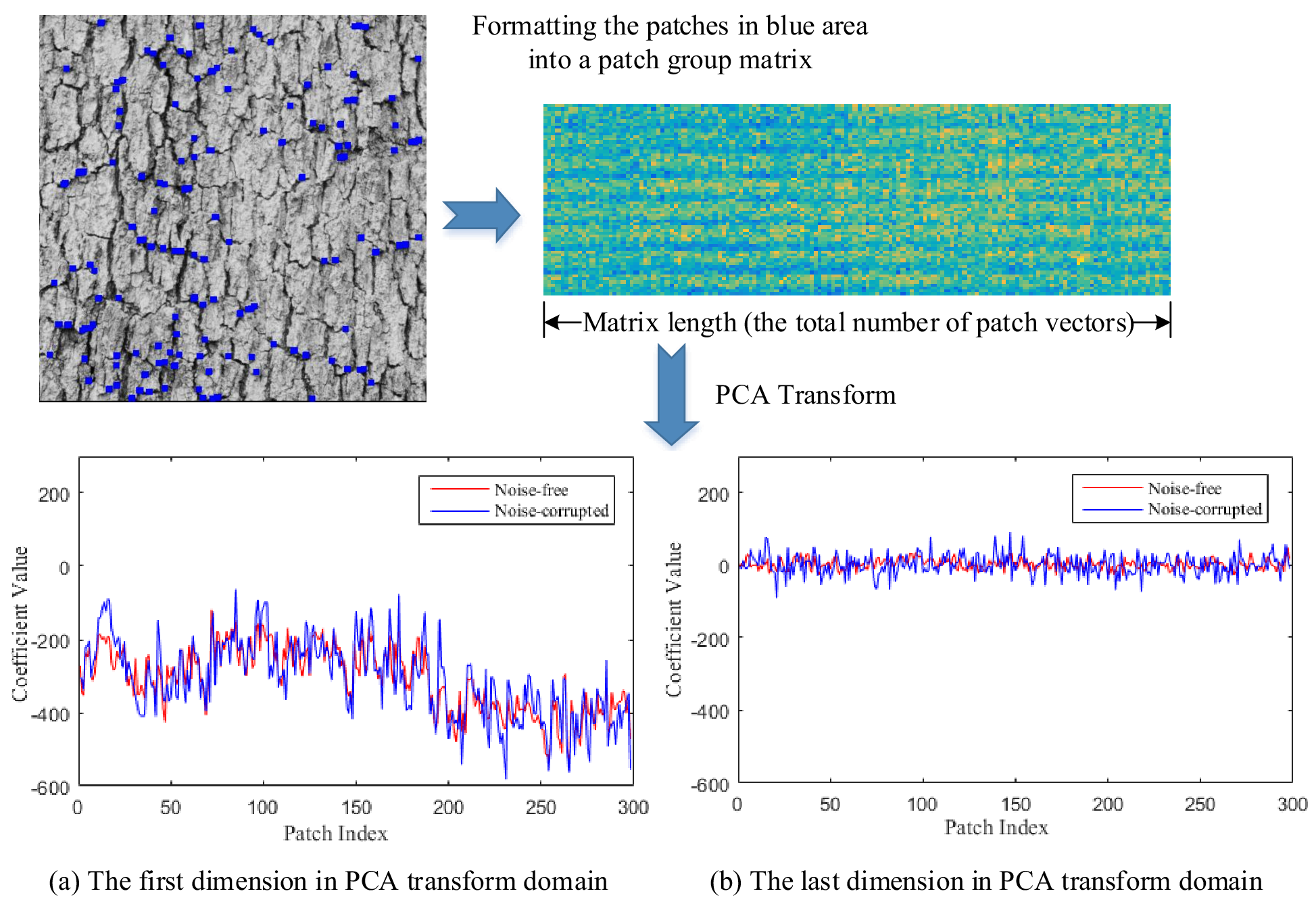}}
 \caption{\label{fig1} The between- and within-dimension variations of PCA transform coefficients for a patch-group matrix consisting of similar patches. (a) The first dimension (signal-dominant) in PCA transform domain, (b) The last dimension (noise-dominant) in PCA transform domain. The difference between (a) and (b) shows the between-dimension variations, while the drastic fluctuation of coefficient value within (a) shows the within-dimension variations.}
\end{figure}

PG techniques collect similar (high-correlated) patches together so that DF can exploit the NSS to boost the denoising performance. During the PG process, if dissimilar patches are gathered in the same patch group, it would be much more difficult for DF to preserve the texture variations. The most widely-used PG techniques are block matching and K-means clustering. Unfortunately, they perform poorly in gathering similar patches under noise interference due to their respective deficiencies.
\begin{enumerate}[(i)]
\item  Block matching method\cite{BM3D}\cite{LPG-PCA}\cite{BM3DSAPCA}\cite{WNNM}\cite{SLRD} is based on the computation of Euclidean distance between patch vectors, which is typically not robust to noise. Moreover, the size of patch groups in block matching is usually set manually so that some dissimilar patches may be grouped together.
\item  K-means clustering algorithm divides the data points of a dataset into a fixed number of clusters such that some distance metric related to the centroids of the clusters is minimized. On the one hand, the optimal cluster number cannot be determined easily\cite{PLOW}\cite{K-LLD}\cite{SGHP}. On the other hand, applying K-means clustering to image patches can lead to heavy computational burden due to the high dimensionality of image data.
\end{enumerate}
To overcome these problems, an efficient adaptive clustering method is designed in AC-PT \cite{acpt}, which not only determines the optimal cluster number automatically, but also accelerates the clustering without dimension reduction that can lead to the information loss. However, in case of high noise level, slight under-segmentation still can be observed.

With similar patches collected by PG, it is important to find a suitable DF for the texture-preserving denoising of the patch groups. For the DF design, the state-of-the-art non-local methods usually incorporate NSS with transform domain methods to decorrelate the dimensions of patch vector so that most of the variations among the highly correlated patches are retained in some of the dimensions, with noise evenly distributed among all the dimensions, thus favors to improve denoising performance. Classical transform methods in DF usually use fixed bases such as discrete wavelet transform, discrete cosine transform. One typical algorithm is BM3D that uses fixed 3D transform to achieve an efficient denoising performance. However, fixed transform bases are not enough to represent the complex natural textures and often brings in artifacts.

Compared with fixed transform, adaptive transform, such as sparse representation and PCA, shows fewer artifacts. We see that the integration of NSS with sparse representation leads to excellent denoising performance, for example, \cite{K-LLD} and \cite{LSSC}. However, stochastic texture variation that behaviors similarly to noise can not be represented sparsely. Thus the optimization based on sparsity prior can lead to the loss of stochastic texture information.

In the past decade, many PCA-based denoising methods have achieved state-of-the-art denoising performance. In the PCA transform domain, the energy of PCA coefficients among different dimensions corresponds to their respective eigenvalues and varies from each other; Meanwhile, within each dimension, especially the first few dimensions with the highest eigenvalues, both the noisy coefficients and their noise-free counterparts show a high variation, even though they come from the similar patches collected via PG (see Fig. 1). This drastic within-dimension variations of the transform coefficients come from the variations in natural image textures. Currently, many PCA-based denoising algorithms only consider the between-dimension energy variations and fail to recognize the within-dimension variations for the texture preservation\cite{LPG-PCA}\cite{WNNM}\cite{clusterPCA}\cite{Guo2016An}. Specifically, the iterative and non-iterative singular value (eigenvalue) thresholding (SVT) algorithms specialize in shrinking the singular values (eigenvalues) of dimensions based on the low rank prior, while PCA transform domain filtering based methods such as \cite{LPG-PCA}, and \cite{clusterPCA} simply employ a "global" Wiener filter whose filter parameters are estimated via a dimension-wise overall averaging. Recently, a novel detail-preserving denoising algorithm, AC-PT is proposed to preserve this within-dimension variations via a combination of two thresholding operation: the hard thresholding of eigenvalues and the PCA-domain Wiener filtering with the filter parameters locally estimated, which are designed based on the consideration of between-dimension energy variations and within-dimension variations, respectively. However, the simple Wiener filter itself may cause over-smoothing and loss of texture detail as demonstrated in \cite{suboptimalwiener}, and the fixed window width with which filter parameters are locally estimated also makes a not robust denoising performance.

In this work, an image denoising algorithm that preserves textures is proposed via PCA-transform-domain texture Variation Adaptive filtering for Adaptive Clustered patches (ACVA). We overcome the deficiencies of abovementioned algorithms and achieve a better denoising performance both quantitatively and visually. Generally, the contributions of this paper are embodied in four aspects:
\begin{itemize}
\item For PG, we improve the robustness of adaptive clustering method to high noise level by using a custom merging threshold (for iterative merging) that is a function of both noise level and cluster size instead of noise level only.
\item For DF, considering the texture variations wrapped in the PCA coefficients of each signal-dominant dimension and the characteristic of over-smoothing of Wiener filter, we perform an improved denoising of the signal-dominant dimensions using the coefficient-wise suboptimal Wiener filter with the filter parameters tracking texture variation adaptively. Specifically, the window width used for local parameter estimation is not fixed as done in ACPT \cite{acpt} but adaptively calculated by local polynomial approximation-intersection of confidence intervals (LPA-ICI) technique.
\item To avoid the significant increasing of computation burden of adaptive clustering along with the image size, we adopt a sliding-window-and-aggregation approach with fixed window size for better denoising performance.
\item Besides of additive Gaussian reduction, the proposed denoising method is applied to removing Poisson-Gaussian noise in the camera raw image.
\end{itemize}

The rest of the paper is organized as follows. In Section 2 we introduce the noise model. Section 3, 4 and 5 are about on the details of the adaptive patch clustering, texture variation adaptive filtering for PCA coefficients and the sliding window and aggregation technique, respectively. Experimental results are displayed in Section 6. Finally, conclusion is given in Section 7.

\section{Noise model}
The additive white Gaussian noise (AWGN) is written as:
\begin{eqnarray}
\label{eqn_mixednoise}
x(\mathbf{c})=y+n,
\end{eqnarray}
where $y$ is noise-free data, $x$ is noisy, and $n$ follows the normal distribution with zero mean and variance $\sigma^2$. AWGN is signal-independent.

Being different from AWGN, the Poisson-Gaussian noise corrupting the camera raw images that are acquired from digital cameras are typically signal-dependent noise. Let $x$ be a noise-free signal at the position $c$. The observed data with Poisson-Gaussian noise can be written as:

\begin{eqnarray}
\label{eqn_mixednoise}
\tilde{x}(\mathbf{c})=\rho/\alpha+bv,
\end{eqnarray}
where $\rho\thicksim P(\alpha(x(\mathbf{c})-p))$ is a Poisson variable with the parameter $\alpha(x(\mathbf{c})-p)$, $v$ follows the normal distribution $N(0,1)$, and $\alpha$, $b$, $p$ are parameters of the Poisson-Gaussian noise.
\begin{figure}[htb]
 \center{\includegraphics[width=9.0cm]  {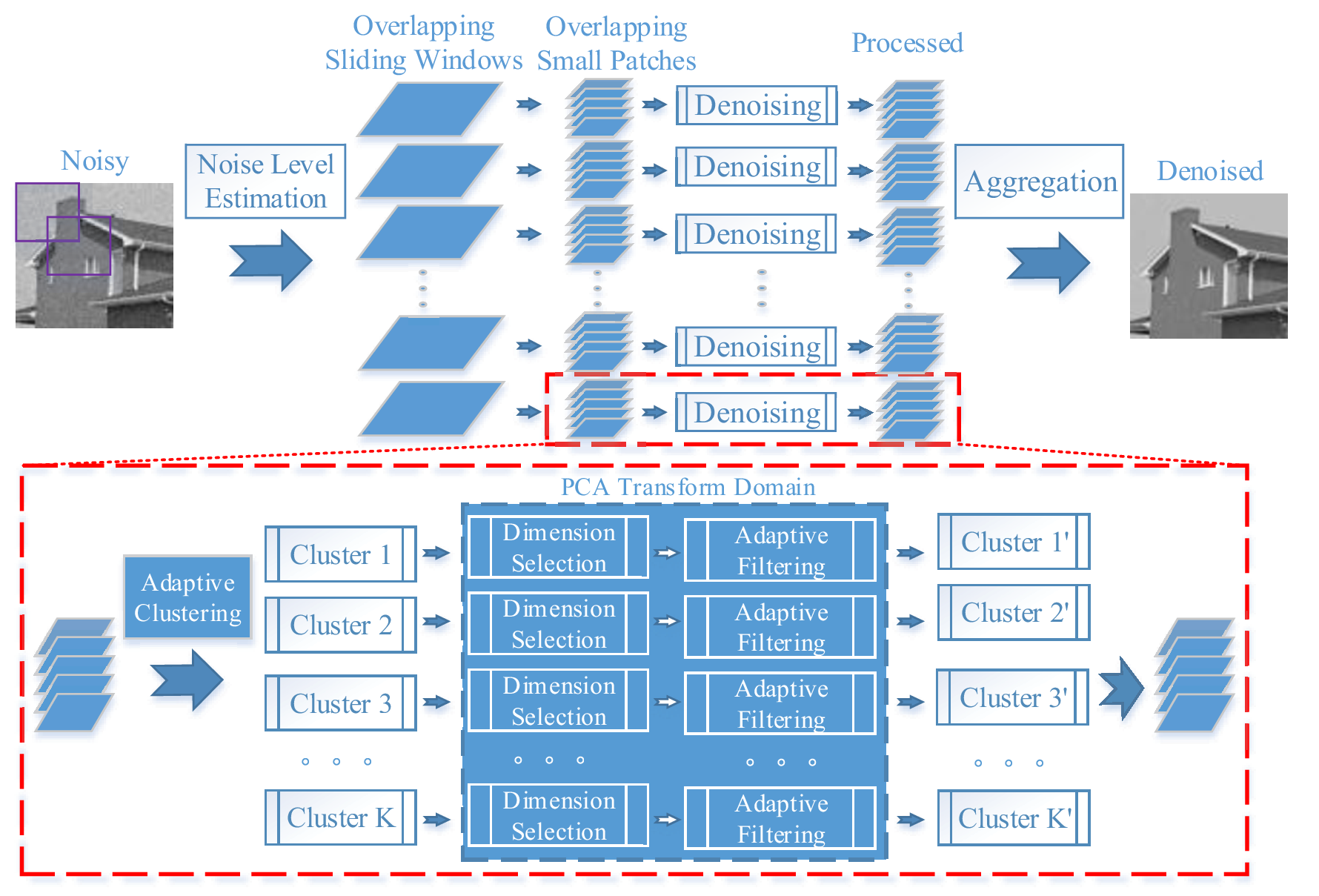}}
 \caption{\label{fig2} Flowchart for the proposed algorithm.}
\end{figure}

After applying a variance stabilization transform for the signal-dependent Poisson-Gaussian noisy signal, we can remove the noise using the denoising methods for additive white Gaussian noise. One well-known variance stabilization transform is called generalized Anscome transform (GAT)\cite{foi2008practical}\cite{makitalo2013optimal}. GAT can approximately transform Poisson-Gaussian noise into additive white Gaussian noise with unitary variance:
\begin{eqnarray}
\label{eqn_GAT}
f(x)=\left\{\begin{array}{lr}
             2\sqrt{x'+\frac{3}{8}+{\sigma'}^2}, &  x'>-\frac{3}{8}-{\sigma'}^2\\
             0, &  x'\leq -\frac{3}{8}-{\sigma'}^2
             \end{array}
\right.
\end{eqnarray}
where $x'=\alpha x$ and $\sigma'=\alpha b$.

Let $y$ be the noise-free data, and the denoised data $y'$ is treated as $E[f(x)|y]$. The exact unbiased inverse of the GAT is defined as:
\begin{eqnarray}
\label{eqn_inverseGAT}
T^{(IGAT)}:E[f(x)|y]\longmapsto E[x|y],
\end{eqnarray}
where $E[x|y]=y$, and $E[f(x)|y]=2\sum_{x=0}^{+\infty}(\sqrt{x+\frac{3}{8}}\cdot\frac{y^x e^{-y}}{x!})$.

A better camera raw image denoising with GAT comes to require a better Gaussian denoising algorithm. Before we detail the proposed denoising algorithms, we want to make clear the symbol system. Generally, we use lower (upper) case bold-face letters to stand for column vectors (matrices). Denote by $\mathbf{X}=(\mathbf{x})\in \mathbb{R}^{M\times L}$ a $M\times L$ matrix with column vectors $\mathbf{x}_i, 1\leq i \leq L$.
Superscript $\ ^T$ represents transpose of a vector or a matrix. Given an image $\mathbf{\Phi}\in \mathbb{R}^{s\times t}$, the total number of all the possible $d\times d$ overlapping patches ${{\mathbf{P}}_{i}}\in {{\mathbb{R}}^{d\times d}}$ is $L=(s-d+1)\times (t-d+1)$ with $i\in \{1,2,\cdots,L\}$. The observation vector $\mathbf{y}_i\in \mathbb{R}^{M\times 1}$ with $M=d^2$ is constructed by stretching the patch $\mathbf{P}_i$. So the image $\Phi$ can be represented as $\mathbf{Y_{\Phi}} \in \mathbb{R}^{M\times L}$ with each column a stretched patch.
A certain cluster can be represented with a matrix with each column a stretched patch. For any data matrix $\mathbf{X}$, we add the subscript $ _c$ to denote the centralized matrix $\mathbf{X}_c=\mathbf{X}-E(\mathbf{X})$, where $E(\cdot)$ represents the expectation.

The proposed denoising method is depicted in Fig. \ref{fig2}. In the noise level estimation step, the noise level can be estimated as in \cite{acpt} for Gaussian denoising. And for Poisson-Gaussian denoising, we can estimate the noise parameters as in \cite{foi2008practical} and then transform Poisson-Gaussian noise into additive white Gaussian noise with unitary variance.
In the following part of this paper, we focus on the illustration of the proposed denoising algorithm: (1) adaptive patch clustering for PG in Section \ref{Sec:AdaptiveClustering}; (2) variation-adaptive filtering in PCA transform domain for DF in Section \ref{Sec:VariationAdaptive}; (3) sliding window and aggregation technique in Section \ref{Sec:SlidingWindow}.

\section{AC step: adaptive clustering of patches}
\label{Sec:AdaptiveClustering}
Many popular clustering algorithms have a common deficiency that the optimal cluster number is difficult to be determined. However, it is easy for us to estimate an approximate range of the cluster number. Supposing the patch size is $d\times d$ and the image size is $s\times t$, in most cases, the maximal cluster number should be below $\frac{st}{d^2}$. Assuming each pixel to be the center of a image patch, we can obtain the maximal cluster number by separating the image into non-overlapping small segments. This small segment with area approximately equal to $d^2$ represents a distinct cluster. Meanwhile, the minimal cluster number is $1$.

Given the approximate range of cluster number, an intuitive idea is that we can determine the optimal cluster number by first obtaining the maximal number of clusters, and then iteratively merging these clusters according to a custom threshold. To this end, there are two problems that need to be solved:
\begin{enumerate}[a)]
\item clustering a huge number of clusters requires a huge computation burden due to the high dimension of image patches;
\item finding a way to calculate a suitable merging threshold.
\end{enumerate}

For the first problem, we adopt the divide and conquer technique. The divide and conquer technique is a two-stage clustering scheme, which accelerate the K-means clustering with improved performance: It first clusters a small number of clusters using K-means, and then within each cluster it performs the K-means clustering again to further increase the cluster number.

For the second problem, we derive the merging threshold on the distance of any two clusters according to the noise level and cluster size. Specifically, we consider one special case, where we have two clusters $\mathbf{A}\in \mathbb{R}^{M\times L_a}$ and $\mathbf{B} \in \mathbb{R}^{M\times L_b}$ with very different sizes $L_a \gg 1$ and $L_b =1$. Denote by $\mathbf{y}_a=\mathbf{y}$ and $\mathbf{y}_b=\mathbf{y}+\mathbf{n}$ the centers of $\mathbf{A}$ and $\mathbf{B}$ respectively, where $\mathbf{y}$ is noise-free, and the entries $n_i\sim N(0,1),1\leq i\leq M$ of vector $\mathbf{n}$ are independent and identically distributed (i. i. d. ). The between-cluster distance is
\begin{equation}
\label{eqn_threshold}
D(\mathbf{B},\mathbf{A})^2=\|\mathbf{y}_B-\mathbf{y}_A\|_2^2=\|\mathbf{n}\|_2^2=\sum_{i=1}^{M}n_i^2,
\end{equation}
and $D(\mathbf{B},\mathbf{A})^2$ follows the chi-squared distribution with $M$ degrees of freedom. These two clusters obtained from K-means with so huge discrepancies in size usually have a very low probability of belonging to the same kind of feature. So the probability we merging the two clusters is $Prob (D(\mathbf{B},\mathbf{A})^2 < \xi)=\varepsilon$, where $\xi$ is the merging threshold and $\varepsilon$ is a very small value. If we set $M=64$ and $\varepsilon= 1.3\times10^{-10}$, we have $\xi \approx 16.0$. Furthermore, if $n_i\sim N(0,\sigma^2),1\leq i\leq M$, $\xi \approx 16.0\sigma^2$. When cluster size of $\mathbf{B}$ become larger, the noise variance in its center $\mathbf{y}_b$ will decrease. Therefore, the merging threshold $\xi$ derived above is essentially the largest acceptable dissimilarity between the two similar clusters that we want to merge together.

Simply using the merging threshold computed according to the noise level $\sigma$ has led to favorable denoising performance as in our previous work\cite{acpt}, but there is still a problem: in case of two large size clusters, the merging threshold is too large because that the influence of noise to the cluster center of the large size clusters are small enough to be ignored. Therefore, when the minimum size of two clusters $L_s$ is larger than a certain value $L_T$, we decrease the possibility of cluster merging by amplifying the between-cluster distance with an amplification coefficient $\rho$: $\hat{D}(\mathbf{B},\mathbf{A})^2=D(\mathbf{B},\mathbf{A})^2/\rho$. We empirically set $L_T=200$ and $\rho=0.7$ to get a satisfactory performance. The comparison of different clustering methods is displayed in Fig. 3. The clustering result based on K-means using its optimal cluster number is somewhat under-segmented, while the segmentation by over-clustering is a typical over-segmentation. By comparison, the two results by adaptive clustering methods is more favorable. Moreover, in the black oval, we see that segmentation based on the improved adaptive clustering reflects more changes on wall texture than its previous version in AC-PT\cite{acpt} without over-segmentation of the sky.

In sum, we conclude the AC-step of the proposed ACVA method in Algorithm \ref{alg:adaptive clustering}.

\begin{algorithm}[htb]
\caption{Adaptive clustering via over-clustering and iterative merging}
\label{alg:adaptive clustering}
\begin{algorithmic}[1]
\Require
 Gaussian noisy data $\Phi\in\mathbb{R}^{s\times t}$, noise level $\sigma$
 \Ensure
Cluster matrices
\State Patch extraction: partition the image into the collection of image patches $X$ with patch size $d=8$
\State Initial clustering: clustering $K^1=\max\{\frac{s\times t}{256\times 256},4\}$ clusters $\mathbf{X}_k, 1\leq k\leq K^1$ with K-means
\For{each cluster $\mathbf{X}_k$}
\State re-clustering $K^2_k\max\{\frac{L_i}{d\times d},1\}$ clusters with K-means
\EndFor
\State Collect all the clusters
\While{the minimum between-cluster distance is larger than threshold $\xi$}
\State Compute between-cluster distance for all the possible cluster pairs and reorder the distance results in ascending order
\If{there exists two clusters whose cluster sizes are larger than $L_T=200$}
\State Amplify the distance with the coefficient $\rho^{-1}$
\EndIf
\State Merge all the cluster pairs whose between-cluster-distance is below $\xi$
\EndWhile 
\end{algorithmic}
\end{algorithm}

\begin{figure}[htb]
 \center{\includegraphics[width=9.0cm]  {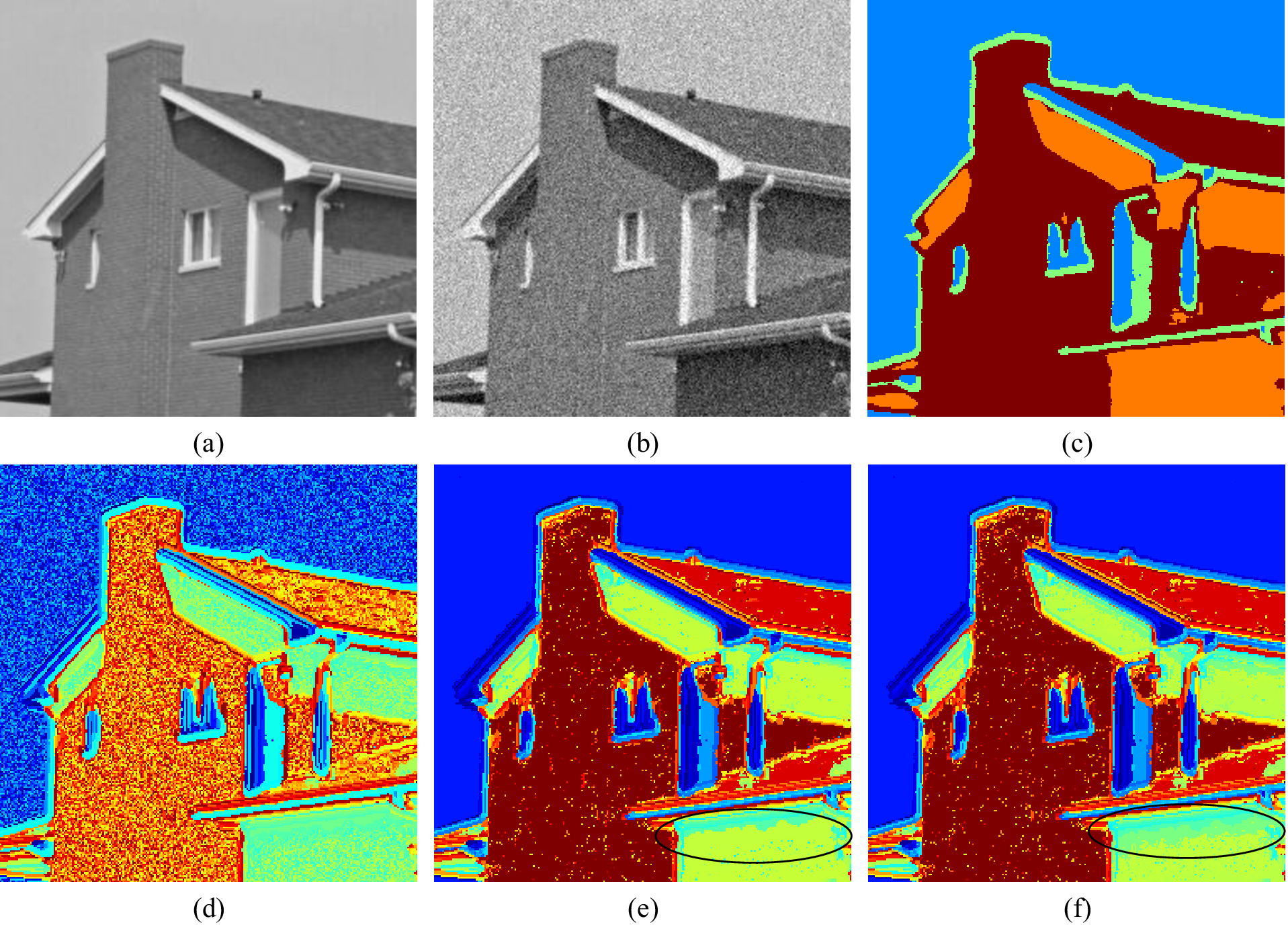}}
 \caption{\label{fig3}The segmentation results on the noisy image with $\sigma=20$ using K-means clustering and the adaptive clustering. (a) House; (b) Noisy image; (c) K-means clustering(optimal number of clusters); (d) Over-clustering based on divide-and-conquer technique; (e)Adaptive clustering as in AC-PT\cite{acpt}; (f) Adaptive clustering considering the cluster size. }
\end{figure}

\section{VA-step: variation-adaptive filtering in PCA domain}
\label{Sec:VariationAdaptive}
In PCA transform domain, inspired by the between-dimension energy variations and the within-dimension variations of PCA coefficients in the signal dominant dimensions with the highest eigenvalues (see Fig. 1), we use a two-step texture variation adaptive approximation strategy to achieve a texture-preserving denoising performance. First, a low rank approximation is implemented via dimension selection based on hard thresholding of eigenvalues to selectively preserve the energy variations of the signal dominant dimensions. Second, each signal dominant dimension is further denoised adaptively via a coefficient-wise adaptive filter with locally estimated filter parameters to protect the underlying within-dimension texture variation.

\subsection{Dimension selection considering the between-dimension energy variations}
Since the texture information is hardly remained in the noise-dominant dimensions with the lowest eigenvalues, we discard the noise-dominant dimensions via dimension selection before the within-dimension filtering to reduce the computational cost and improve the denoising performance.

Considering the centralized noisy cluster matrix $\mathbf{X}_c=(\mathbf{x}_c)\in \mathbb{R}^{M\times L}$, $\mathbf{X}_c=\mathbf{X}_{c,0}+\mathbf{N}$, where $\mathbf{X}_{c,0}=(\mathbf{x}_{c,0})\in \mathbb{R}^{M\times L}$ is noise-free and $\mathbf{N}=(\mathbf{n})\in \mathbb{R}^{M\times L}$ is the noise matrix with each column vector $\mathbf{n}_i\sim N_M(0,\sigma^2 \mathbf{I})$, where $\mathbf{I}$ is identity matrix. Suppose $\mathbf{X}_{c}=\sqrt{L}\sum_{i=1}^{\min(M,L))}\sqrt{\lambda_i}\mathbf{u}_i\mathbf{v}_i^T$ and $\mathbf{X}_{c,0}= \sqrt{L}\sum_{i=1}^{R}\sqrt{\lambda_{i,0}}\mathbf{u}_{i,0}\mathbf{v}_{i,0}^T$ with singular values $\sqrt{L\lambda_i}$ and $\sqrt{L\lambda_{i,0}}$, and singular vectors $\mathbf{u}_i$,$\mathbf{v}_i$, $\mathbf{u}_{i,0}$ and $\mathbf{v}_{i,0}$, $1\leq i\leq \min(M,L)$.

The dimension selection can be done based on the Gaussian spiked population model \cite{GSURE}. Suppose $\gamma=L/M$ is a constant. Letting $L\rightarrow \infty$ and $\lambda_{n\pm}=\sigma^2(1\pm\sqrt{\gamma})^2$, there is:
\begin{eqnarray}
\label{eqn_select}
\lim_{L\rightarrow\infty} \lambda_i=\left\{
\begin{array}{lll}
\rho(\lambda_{i,0}), \ \ if\ i\leq R\ and\  \lambda_{i,0}>\sigma^2\gamma^{1/2}. \\
\lambda_{n+}, \ \ otherwise.
\end{array}
\right.
\end{eqnarray}
where $\rho$ is a real-valued function.

The Gaussian spiked population model implies that for the normal size cluster matrices the eigenvalues of noise-dominant dimensions are below $\lambda_{n+}$ approximately. Since the dimensions with eigenvalues close to $\lambda_{n+}$ are still noisy, we set a correction coefficient $\mu$ to estimate the rank $R$ as:
\begin{eqnarray}
\label{eqn_tildeX}
R\approx\sum_{i=1}^{M}\mathbf{1}(\lambda_{i}>\mu\lambda_{n+})
\end{eqnarray}

Thus the low rank approximation can be computed as:
\begin{eqnarray}
\label{eqn_tildeX}
\mathbf{X}_R=\sqrt{L}\sum_{i=1}^{R}\sqrt{\lambda_{i}}\mathbf{u}_{x,i}\mathbf{v}_{x,i}^T.
\end{eqnarray}
Similar to \cite{acpt}, we set $\mu=1.1$ to get a favorable denoising performance.

\subsection{Within-dimension variation adaptive filtering}
Consider the low-rank matrix $\mathbf{X}_R=\mathbf{U}_R \mathbf{P}_R$ obtained in the previous section, where $\mathbf{U}_R$ consists of the selected eigenvectors $\mathbf{U}_R= [\mathbf{u}_{x,1},\mathbf{u}_{x,2}, \cdots, \mathbf{u}_{x,R}]$ , and $\mathbf{P}_R$ consists of the corresponding signal-dominant dimensions in PCA transform domain:
\begin{equation}
\label{eqn_transform}
\mathbf{P}_R = [\mathbf{p}_1 ,\mathbf{p}_2 , \cdots, \mathbf{p}_R]^T,
\end{equation}
and $\mathbf{p}_i= \sqrt{L\lambda_{i}} \mathbf{v}_{x,i}$ ($1\leq i\leq R$) is the selected PCA dimension.

For illustrative purpose, we further extract the coefficients in any dimension $\mathbf{p}_i=[p_{i,1},p_{i,2},\cdots,p_{i,L}]$ and denote these coefficients in the $i$th dimension as the noisy observations of a "signal sequence" containing $L$ observation points: $y(n)=p_{i,n}$$,$ $n=1,2, \cdots, L$. Let $y(n)=f(n)+w(n)$, where $w(n)$ is i.d.d Gaussian noise of zero mean and variance $\sigma_w^2 = \sigma^2$, and $f(n)$ (with variance $\sigma_f^2$) is the noise-free signal that we want to estimate. Thus for an observation at a point $n$, there is $y(n)\sim N(f(n),\sigma^2)$. Here, some specific observations are considered as being "similar" to each other when their respective means $f(n)$ are close to each other, and we say that there is a considerable "variation" between the observations when their respective means are different from each other.

Before we detail how to denoise each dimension $y(n)$ and obtain the estimate $\hat{f}(n)$ for each dimension, we must learn some useful characteristics on the extracted signal sequence $y(n)$ and its corresponding $f(n)$. One typical example of this kind of the signal sequence is displayed in Fig 1. (a), where the vertical axis corresponds to the signal value for the noisy signal $y(n)$ (blue line) and noise-free signal $f(n)$ (red line) and the horizonal axis is the sequence index $n$ corresponding to different patches. We see that the signal values are aggregated densely in a certain interval, implying that their respective means are close to each other. This dense aggregation of all the means can be regarded as a global similarity between all the observations. Moreover, there is also a drastic and irregular fluctuation for both the noise-corrupted signal $y(n)$ and noise-free signal $f(n)$. This fluctuation of noise-free signal (i.e. the sequence of the means) is called the internal variation of the signal sequence.
The global similarity and internal variation of the signal sequence in PCA transform domain can be interpreted in the corresponding spatial domain. Every transform-domain signal observation comes from a certain patch of specific patch group in spatial domain. All the pathes in the same patch group are similar to each other (i.e. a global similarity for the whole patch group), while the patches within the patch group are still different from each other to some extent (i.e. an internal variation). Both the global similarity and the internal variations in spatial domain are transformed into the orthogonal PCA transform domain.

Considering the above characteristics in estimating each signal observation for PCA coefficient, we have three aspects of consideration: 1) The noise-free signal is rich in unsmooth and irregular variations that convey rich texture information. These unsmooth and irregular variations will be distorted by the denoising filters based on smoothness (continuity) constraints; 2) Based on the global similarity between the signal observations, we can find the sample observations with high similarity to the signal to be estimated and average them to get an optimal estimate for boosting denoising performance; 3) To preserve the internal variations within the observations, we must choose the sample observations adaptively according the signal to be estimated, such that the high similarity observations around the signal to be estimated should be selected while other dissimilar observations need to be excluded.

For the consideration 1), we estimate $f(n)$ by applying an efficient Wiener filter that has been used for the favorable PCA domain filtering by many denoising algorithms, and we particularly design a suboptimal Wiener filtering to avoid signal distortion and preserve the variations. However, the satisfactory variation-preserving performance of suboptimal Wiener filter is highly dependent on an optimal estimate of its parameter, i.e. the auto-covariance of $y(n)$: ${R}_{y}(n)=E(y(n)^2)$. Based on the consideration 2) and 3), we estimate the auto-covariance of $y(n)$ in a point-wise approach. For each signal its auto-covariance is estimated via an adaptive local average of similar observations. As a whole, this point-wise suboptimal Wiener filtering consists of two steps: locally adaptive filter parameter estimation and suboptimal Wiener filtering using the estimated filter parameter.

\subsubsection{Estimating the filter parameter locally based on LPA-ICI}
To denoise the signal observation $y(n)$ at point $n$ with Wiener filter, we need to estimate its filter parameter, i.e., auto-covariance ${R}_{y}(n)$ to further compute the cross-covariance ${R}_{fy}(n)={R}_{y}(n)-\sigma^2$ between $y(n)$ and its noise-free counterpart $f(n)$. The auto-covariance at point $n$ can be written as: ${R}_{y}(n)= E(y(n)^2)=f(n)^2+\sigma^2$, where the expected value can be approximated statistically via an average of several squared sample signal observations.

The remaining problem is how to accurately select the signal observations with high similarity to the signal (at point $n$) to be estimated. A simple practical solution is to select the observations in a local neighborhood of the signal point $n$ because relatively higher similarity is more likely to happen in the local neighborhood due to the property of the applied PG technique. In particular, for AC, a large cluster matrix consists of many small local segments that are actually small clusters matrices generated from the over-clustering stage in AC and typically have a higher similarity.
Furthermore, in \cite{acpt}, the local neighborhood has been used for the filter parameter estimate and proves to achieve a better denoising performance than the global approach.

Being different from \cite{acpt} that chooses the local neighborhood in a fixed-size window for local estimate, the proposed method use LPA-ICI to adaptively determine the window width for a better local estimate. Here, choosing the window width is essentially equivalent to choosing the estimate samples. A small width corresponds to a smaller moving window with fewer samples for the local parameter estimate and therefore to noisier estimates, with higher variance and typically decreased estimation bias, and vice versa. Therefore, the window width controls the trade-off between the bias and variance in the local estimate and the varying window width for the local estimate is very important. This assumption is confirmed by the fact that in density estimation and signal reconstruction studies in the literature, almost all the adaptive-width windows \cite{bandwidth2014}\cite{Qin2018} have been shown to be superior to fixed-width windows.

The local polynomial approximation (LPA) method, combined with intersection of confidence intervals (ICI) rule is a method originally developed for pointwise adaptive estimation of $1$-D signal \cite{LPAICI02}. In this work we only use it for the purpose of detecting variations and finding similar samples. The use of LPA-ICI for variations detection has been used in some image denoising algorithms such as SADCT \cite{shapeadaptiveDCT} and BM3DSAPCA \cite{BM3DSAPCA}, to adaptively detect the spatial variations of image value and collect similar pixel samples. However, the proposed algorithm uses LPA-ICI for signal variation detection in the PCA transform-domain.

Standard linear LPA tries to fit the signal $y(n)$ locally with polynomial functions of order $m$. Here, since we only use it to detect variations and find neighborhood with high internal similarity, we simply apply the zero-order polynomial fitting ($m=0$) to find a suitable window of size $h$ (a window containing $N_h=2h+1$ data points) where all the similar signal in the window can be approximated by a constant amplitude signal $\hat{y}(n,h)=C$. The computation of $\hat{y}(n,h)$ in LPA is related to the following loss function:
\begin{eqnarray}
\label{eqn_LPA}
\mathfrak{J}_h(n)=\frac{1}{N_h}\sum_{s=1}^{N_h}\rho_h (n_s-n)(y(n_s)-\hat{y}(n,h))^2
\end{eqnarray}
where $y(n_s), 1\leq s\leq N_h$ is the signal at the point in a window of size $h$ with $n$ its center.
$\rho(\cdot)$ is a basic window function, and $\rho_h(\cdot)=\rho(\cdot/h)/h$. For simplicity, we use the square uniform window, where $\rho(\cdot)=1$ in $[-1,1]$, and $\rho(\cdot)=0$, otherwise. So there is $\rho_h(\cdot)=1/h$ in $[-h,h]$, and  $\rho(\cdot)=0$, otherwise.

By minimizing the loss function, we have the estimate of $y(n)$: $\hat{y}(n,h)=\frac{1}{N_h}\sum_{s=1}^{N_h}y(n_s)$ and its standard deviation ${std}(n,h)=\frac{\sigma}{\sqrt{N_h}}$. So the  confidence interval of the estimate can be
\begin{eqnarray}
\label{eqn_interval}
D=[L,U]\notag\\
U=\hat{y}(n,h)+\Gamma\cdot {std}(n,h)\notag\\
L=\hat{y}(n,h)-\Gamma\cdot {std}(n,h)
\end{eqnarray}
where $\Gamma$ is a threshold parameter.

Given a finite set of window size $H={h_1<h_2<\cdots<h_J}$ starting from the minimum window size $h_1$, for each window we can use the LPA to get a estimate $\hat{y}(n,h_i)$ and a corresponding standard deviation ${std}(n,h_i)$, thereby determining a sequence of the confidence intervals $\mathcal{D}(i)$, $1\leq i\leq J$ of the biased estimates:

\begin{eqnarray}
\label{eqn_intervalsin}
D(i)=[L_i,U_i]\notag\\
U_i=\hat{y}(n,h_i)+\Gamma\cdot {std}(n,h_i)\notag\\
L_i=\hat{y}(n,h_i)-\Gamma\cdot {std}(n,h_i)
\end{eqnarray}

The ICI technique considers the optimal $h$ to be the maximum window length satisfying $\underline{L}_i<\overline{U}_i$, where $\underline{L}_i=\max\{L_i, \underline{L}_{i-1} \}$ and $\overline{U}_i=\min\{L_i, \overline{U}_{i-1} \}$, $1<i\leq J$.

After determining a suitable window size $h$ with $n$ its center, we have the segment $y (n_s)=f(n_s)+w(n_s), 1\leq s\leq N_h$. The estimated auto-covariance of $y(n)$ at point $n$ can be computed via a local average of this segment:
\begin{equation}
\label{eqn_wiener}
{R}_{y}(n)= \frac{1}{N_h}\sum_{s=1}^{N_h}y(n_s)^2.
\end{equation}

\subsubsection{Suboptimal Wiener filtering}
Wiener filter has been widely used to remove the noise in transform domain efficiently\cite{BM3D}\cite{LPG-PCA}.
Given the estimated auto-covariance ${R}_{y}(n)$ at point $n$ and auto-covariance of noise ${R}_{w}=\sigma^2$, the estimate of the observation $f(n)$ at $n$ by Wiener filter is:
\begin{equation}
\label{eqn_wiener}
\hat{f}(n)= {h}_o {y}(n),
\end{equation}
where ${h}_o={R_{y}(n)}^{-1}R_{fy}(n)=[1-g_o]$, and $g_o={R_{y}(n)}^{-1}R_{w}$.

However, as has been observed in \cite{suboptimalwiener}, the optimal Wiener filter often results in signal distortion during the noise reduction.
It is necessary for us to avoid the signal distortion as much as possible, while reducing most of the noise, thereby achieving a satisfactory denoising performance both visually and quantitatively. To achieve a better control of noise reduction and signal distortion, instead of using the optimal Wiener filter, we can use a suboptimal Wiener filter by manipulating the Wiener filter properly and automatically with an attenuation coefficient $\alpha$ of $g_o$ as in \cite{suboptimalwiener}:
\begin{equation}
\label{eqn_suboptimal wiener}
{h}_s=[1-\alpha g_o],
\end{equation}
where $\alpha\in [0,1]$ and $\alpha=0$ and $1$ correspond to the case of identity filter and Wiener filter, respectively. For simplicity, let $g_s=\alpha g_o$. Then the estimate of the observation $f(n)$ at $n$ by the suboptimal Wiener filter is:
\begin{equation}
\label{eqn_suboptimal wiener}
\hat{f}(n)= {h}_s {y}(n),
\end{equation}

The determination of $\alpha$ is based on two indexes relating to signal distortion and noise reduction, respectively. For the suboptimal Wiener filter corresponding to a certain $\alpha$, the signal distortion index can be defined as $v_{sd}(g_s)\triangleq \frac{E\{[f(n)-h_s f(n)]^2\}}{\sigma_f^2}$, and the noise reduction index is ${\xi}_{nr}(h_s)\triangleq \frac{\sigma_w^2}{E\{[h_s w(n)]^2\}}$.

Then, we can obtain the optimal $\alpha$ by maximizing the following discriminative cost function relating noise reduction index and signal distortion index\cite{suboptimalwiener},
\begin{eqnarray}
\label{eqn_alpha}
J(\alpha)&\triangleq &\frac{{\xi}_{nr}(h_s)}{{\xi}_{nr}(h_o)}-\beta \frac{v_{sd}(g_s)}{v_{sd}(g_o)}\\
&=& \frac{\sigma^2+{g_o} {R}_{w} {g_o}-2 \sigma^2 {g_o}}{\sigma^2+\alpha^2 {g_o}{R}_{w} {g_o}-2\alpha \sigma^2 {g_o}}-\beta \alpha^2 ,\notag
\end{eqnarray}
where $\beta$ is an application-dependent constant and determines the relative importance between the improvement in signal distortion and degradation in noise reduction. When $\beta$ becomes larger, we have less signal distortion with less noise removal. We set $\beta=0.7$ as in \cite{suboptimalwiener} to achieve a good balance.

With the suboptimal Wiener filter above, we tackle each $\mathbf{p}_i$ from $\mathbf{P}_R$ and obtain the corresponding processed result $\hat{\mathbf{p}}_i$$,$ $i=1,2,\cdots R$. Then we have $\hat{\mathbf{P}}_R=[\hat{\mathbf{p}}_1,\hat{\mathbf{p}}_2,\cdots, \hat{\mathbf{p}}_R]^T$.
We can further obtain the denoised cluster using the reverse PCA transform:
\begin{equation}
\label{eqn_denoised cluster}
\hat{\mathbf{X}}_R =\mathbf{U}_R \hat{\mathbf{P}}_R.
\end{equation}

\section{Overall of ACVA}
\label{Sec:SlidingWindow}
Considering the high dimensionality of image data matrix, we use a sliding window approach to avoid a significant increase in the computational burden when implementing the adaptive clustering for the image of increasing size. By using a fixed-size sliding window, the computational burden of the proposed algorithm within each window is comparatively stabilized at a reasonable interval. For a sliding window with fixed size $128\times 128$, the runtime of the Matlab codes on a PC equipped with an Intel Core i5-4460 Quad-Core 3.2 GHz CPU ranges from $4.0s$ to $16.0s$ approximately for different noise level.

As shown in Fig. \ref{fig2}, all the estimated patches from the sliding windows at different portions of the image are aggregated and averaged to obtain the final estimate. This sliding window and aggregation approach has also been used in most block matching based denoising algorithms and proves to be helpful to further remove the residual noise in the estimated patches thereby leading to the performance boost.

\begin{algorithm}[htb]
\caption{Variation adaptive filtering based on adaptively clustered patches(ACVA)}
\label{alg:Framwork}
\begin{algorithmic}[1]
\Require
 Gaussian noisy data $\Phi$, noise level $\sigma$.
 \Ensure
The denoised result $\widehat{\Phi}$;
\label{code:fram:extract}
\State Patch extraction: slide a window of size $W_s=128$, and partition the whole block in the window at each portion into a collection of image patches $X$ with patch size $d=8$.
\label{code:fram:trainbase}
\State Adaptive Clustering in Algorithm 1;
\label{code:fram:add}
\For{each cluster}
\State PCA transform;
\State Dimension selection with a hard threshold of PCA eigenvalues;
\For{each selected dimension}
\For{each coefficient}
\State Estimating the parameter of suboptimal Wiener filter with LPA-ICI;
\State Suboptimal Wiener filter denoising;
\EndFor
\EndFor
\State Reverse PCA transform;
\EndFor
\State Reproject all the estimated stacked patches $\widehat{X}$ into image $\widehat{\Phi}$.
\end{algorithmic}
\end{algorithm}
Overall of the proposed algorithm ACVA is summarized in Algorithm \ref{alg:Framwork}.

\begin{figure}[htb]
 \center{\includegraphics[width=9.0cm]  {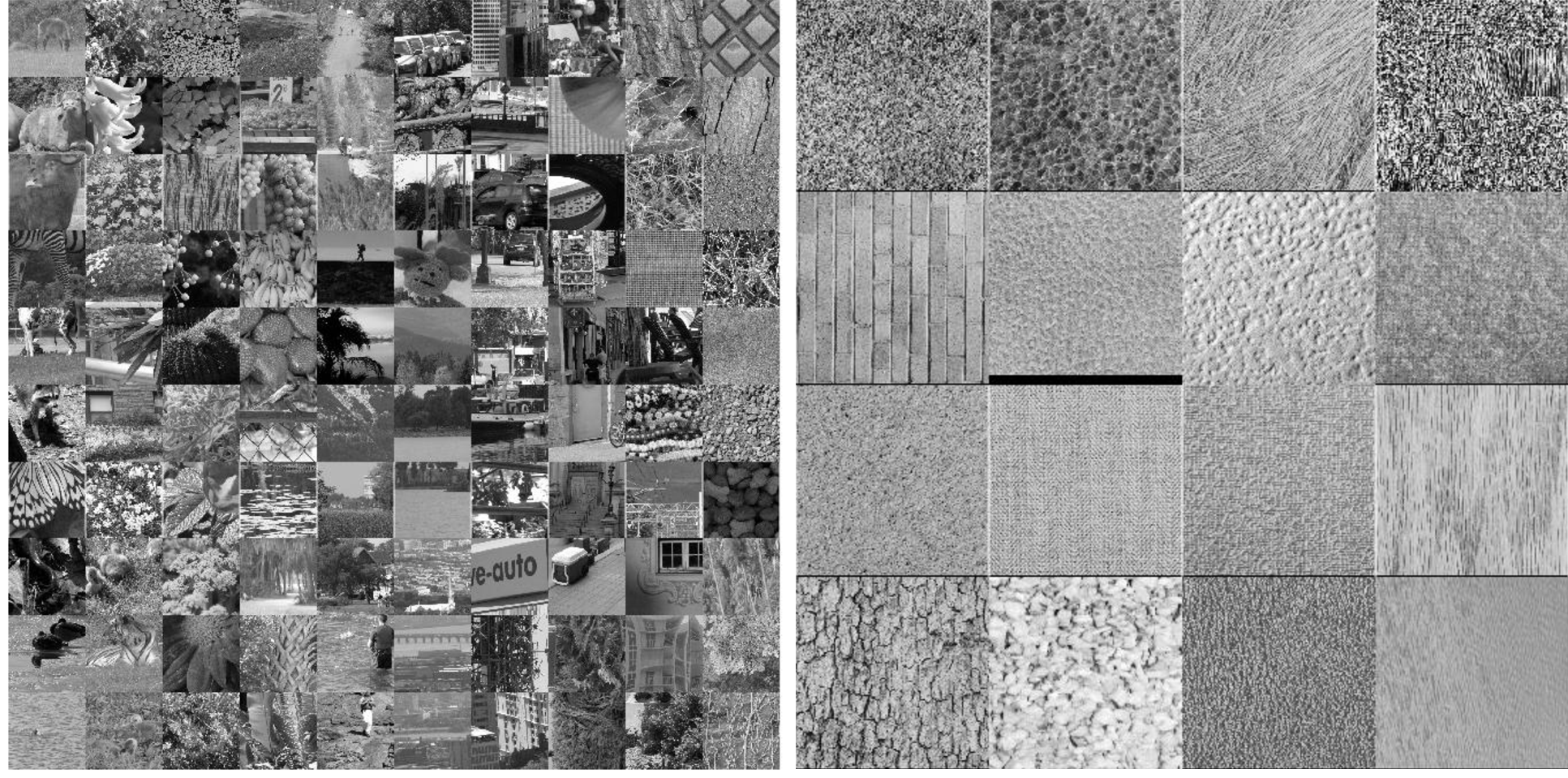}}
 \caption{\label{fig4} The test datasets for Gaussian denoising experiments. Left:The 100 image samples from McGill dataset\cite{McGill} (transformed from RGB into gray images); Right: The 16 texture images from USC-SIPI dataset.}
\end{figure}

\begin{figure}[htb]
 \center{\includegraphics[width=9.0cm]  {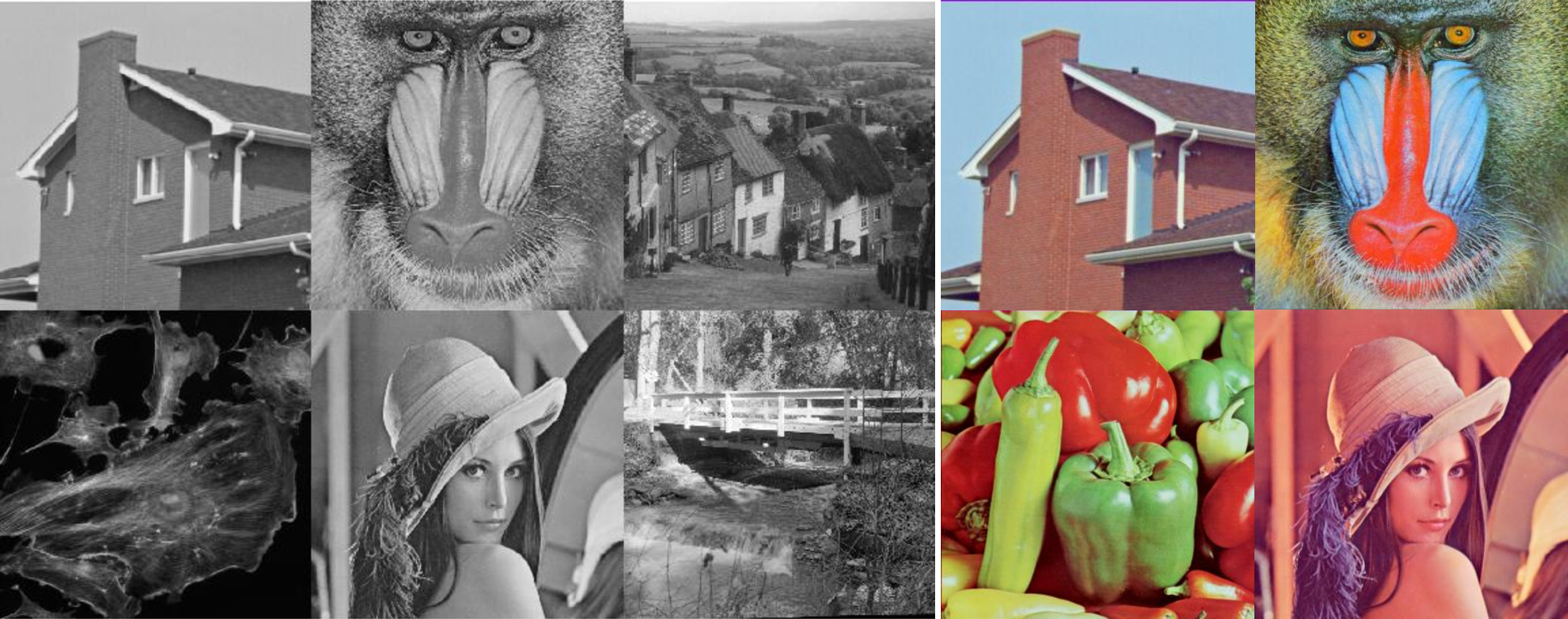}}
 \caption{\label{fig5} The standard test images. Left: six standard gray scale image for Gaussian denoising experiments; Right: four standard RGB image for camera raw image simulation.}
\end{figure}

\section{Experimental results}
To validate all the algorithms comprehensively, we use three performance metrics: peak signal-to-noise ratio(PSNR)\cite{PSNR}, structural similarity(SSIM)\cite{SSIM} and feature similarity(FSIM)\cite{FSIM}. The PSNR regards the structural information  and the nonstructural information as the same in terms of the contribution towards the performance, whereas SSIM and FSIM put more emphasis on the structural information. The images used for denoising experiments are displayed in Fig. \ref{fig4} and \ref{fig5}. We compare the proposed algorithm ACVA\footnote{The Matlab software is available at \url{http://www.escience.cn/people/bjqin/research.html}.} with: BM3D \cite{BM3D}, BM3DSAPCA (SAPCA for short) \cite{BM3DSAPCA}, WNNM\cite{WNNM}, SLRD\cite{SLRD}, SGHP\cite{SGHP}, and AC-PT\cite{acpt}. Moreover, the most representative deep learning based denoising methods, i.e. the denoising convolutional neural network with noise level specified (DnCNN-S)\cite{Zhang2017Beyond}, is also taken into consideration for comparison. All the algorithms are set with default parameters for their best performances.

\subsection{Gaussian denoising}

A) Test on the adaptive clustering: Table \ref{tabe AC} shows how the parameter $L_T$ and $\rho$ in the proposed adaptive clustering affect the denoising performance on $100$ images from the McGill dataset\cite{McGill} at different noise level, where the DF described in the previous section is used. Denote the proposed adaptive clustering method as $AC(L_T,\rho)$, where $L_T$ is the maximum cluster size in a pair of cluster for computation of between-cluster distance and $\rho$ is the amplification efficient. Particularly, the adaptive clustering in AC-PT\cite{acpt} can be denoted as $AC(0,1)$. When setting $L_T\geq 100$ and $\rho\leq0.9$, we see the increase of PSNR especially at high noise level. However, there is no significant difference between the quantitative performance for $\rho=0.9, 0.7$ and $0.5$ in terms of PSNR, SSIM and FSIM. Thus, it is enough to set $L_T=200$ and $\rho=0.7$ for a satisfactory denoising performance.

\begin{figure}[htb]
 \center{\includegraphics[width=8.8cm]  {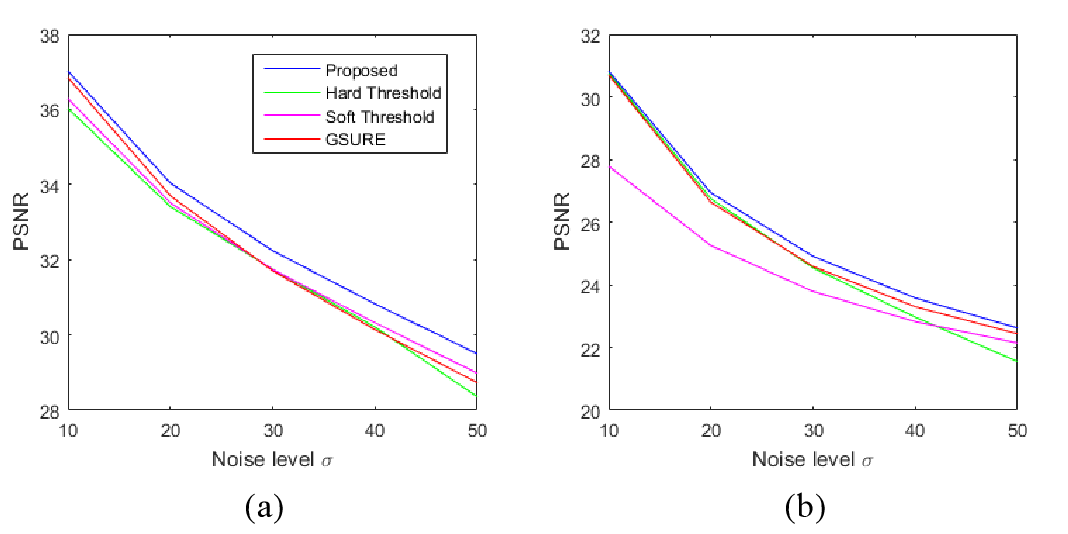}}
 \caption{\label{fig6} PSNR (dB) results of the proposed filter compared with singular value thresholding methods at various noise levels $\sigma$. (a) and (b) are results for House and Bark respectively.}
\end{figure}

\begin{table*}[!t]
\scriptsize
\renewcommand{\arraystretch}{1.0}
\caption{ The average denoising performance of ACVA using Adaptive clustering with different parameters.}
\label{tabe AC}
\centering
\begin{tabular}{|c|c|c|c|c|c|c|c|c|c|}
\hline
$\sigma$&Index&AC(0,0.7)&AC(0,1.0)&AC(100,0.7)&AC(200,0.9)&AC(200,0.7)&AC(200,0.5)&AC(200,0.3)&AC(400,0.7)\\\hline
\multirow{3}{*}{10}&PSNR&32.50&32.50&32.50&32.50&32.50&32.50&32.50&32.50\\
&SSIM&0.9195&0.9196&0.9196&0.9196&0.9196&0.9196&0.9197&0.9196\\
&FSIM&0.9564&0.9568&0.9566&0.9567&0.9567&0.9566&0.9566&0.9566\\\hline
\multirow{3}{*}{20}&PSNR&28.88&28.91&28.91&28.91&28.91&28.91&28.91&28.91\\
&SSIM&0.8383&0.8409&0.8409&0.8409&0.8409&0.8410&0.8410&0.8409\\
&FSIM&0.9141&0.9158&0.9152&0.9154&0.9153&0.9152&0.9150&0.9154\\\hline
\multirow{3}{*}{50}&PSNR&24.60&24.79&24.81&24.80&24.81&24.81&24.81&24.80\\
&SSIM&0.6516&0.6809&0.6810&0.6810&0.6812&0.6810&0.6809&0.6809\\
&FSIM&0.8310&0.8375&0.8362&0.8371&0.8364&0.8359&0.8348&0.8367\\\hline
\multirow{3}{*}{100}&PSNR&21.66&22.05&22.08&22.06&22.09&22.09&22.11&22.07\\
&SSIM&0.4803&0.5347&0.5362&0.5345&0.5367&0.5363&0.5369&0.5351\\
&FSIM&0.7641&0.7561&0.7551&0.7555&0.7555&0.7548&0.7541&0.7552\\\hline
\end{tabular}
\end{table*}

B) Test on the variation adaptive denoising filter:With the adaptive clustering parameters fixed, we further test the proposed texture variation adaptive filter with different settings on the $100$ image samples from McGill dataset. Denote the texture variation adaptive filter as $VA(A,B)$, where $A=W$ for Wiener filter or $A=SW$ for suboptimal Wiener filter ,and $B=N$ for fixed window width or $B=L$ for adaptive window width based on LPA-ICI. Particulary, the setting $VA(W,N)$ corresponds to the DF in AC-PT\cite{acpt}. Table \ref{tabe PA} shows the superior performance of $VA(SW,L)$ to other settings. It should be noted that even through we take a suboptimal approach, the boost of PSNR performance can still be observed, which may imply the characteristics of over-denoising by the traditional optimal Wiener filter.

\begin{table}[!t]
\scriptsize
\renewcommand{\arraystretch}{1.0}
\caption{ The average denoising performance of ACVA using texture variation adaptive filter with different settings.}
\label{tabe PA}
\centering
\begin{tabular}{|c|c|c|c|c|c|c|c|}
\hline
$\sigma$&Index&VA(W,N)&VA(W,L)&VA(SW,N)&VA(SW,L)\\\hline
\multirow{3}{*}{10}&PSNR&32.50&32.49&32.50&32.50\\
&SSIM&0.9179&0.9176&0.9195&0.9196\\
&FSIM&0.9556&0.9553&0.9567&0.9567\\\hline
\multirow{3}{*}{20}&PSNR&28.87&28.85&28.90&28.91\\
&SSIM&0.8368&0.8361&0.8402&0.8409\\
&FSIM&0.9121&0.9109&0.9156&0.9153\\\hline
\multirow{3}{*}{50}&PSNR&24.75&24.73&24.73&24.81\\
&SSIM&0.6767&0.6730&0.6807&0.6812\\
&FSIM&0.8334&0.8268&0.8407&0.8364\\\hline
\multirow{3}{*}{100}&PSNR&22.01&22.00&21.96&22.09\\
&SSIM&0.5321&0.5283&0.5351&0.5367\\
&FSIM&0.7540&0.7421&0.7642&0.7555\\\hline
\end{tabular}
\end{table}

We compare the proposed DF with noniterative SVT methods in recent literature: GSURE\cite{GSURE}, hard thresholding method\cite{hardthrehold-svd}, and soft thresholding method \cite{optimalshrinkSVD16}. As in Fig. \ref{fig6}, our results show that the proposed DF is superior to these noniterative SVT algorithms in denoising Bark and House image.

C) Comparison with the-state-of-the-art denoising algorithms: The images used for test include $16$ textured images(as in Fig. \ref{fig4}) from the USC-SIPI Image Database\footnote{\url{http://sipi.usc.edu/database/}}, six standard test images (Fig. \ref{fig5}), and $100$ image samples from McGill dataset(as in Fig. \ref{fig4}, already transformed into grayscale images). We test the considered algorithms at different noise level $\sigma = 10, 20, 30, 40, 50$.

Table \ref{tabe texture} shows that the proposed method ACVA quantitatively outperforms the other methods in denoising images with irregular textures in terms of PSNR, SSIM and FSIM.

Table \ref{tabe stdimage0} further affirms the competitive quantitative performance, especially in terms of FSIM, to other algorithms on denoising natural images.%

\begin{figure*}[htb]
 \center{\includegraphics[width=18.0cm]  {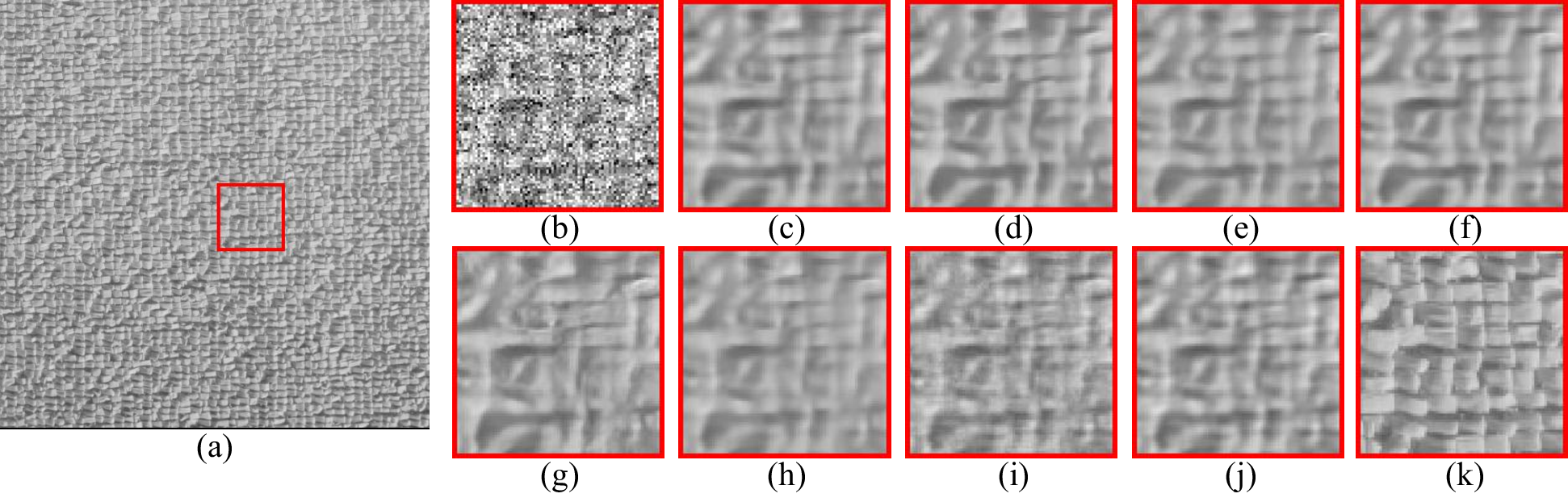}}
 \caption{\label{fig7} Denoising the Grid image at $\sigma =50$:(a)Grid image, (b)Noisy block, (c)BM3D, (d)BM3DSAPCA, (e)WNNM, (f)SLRD, (g)DnCNN-S, (h)SGHP, (i)AC-PT, (j)ACVA, (k)Noise-free block.}
\end{figure*}

\begin{figure*}[htb]
 \center{\includegraphics[width=18.0cm]  {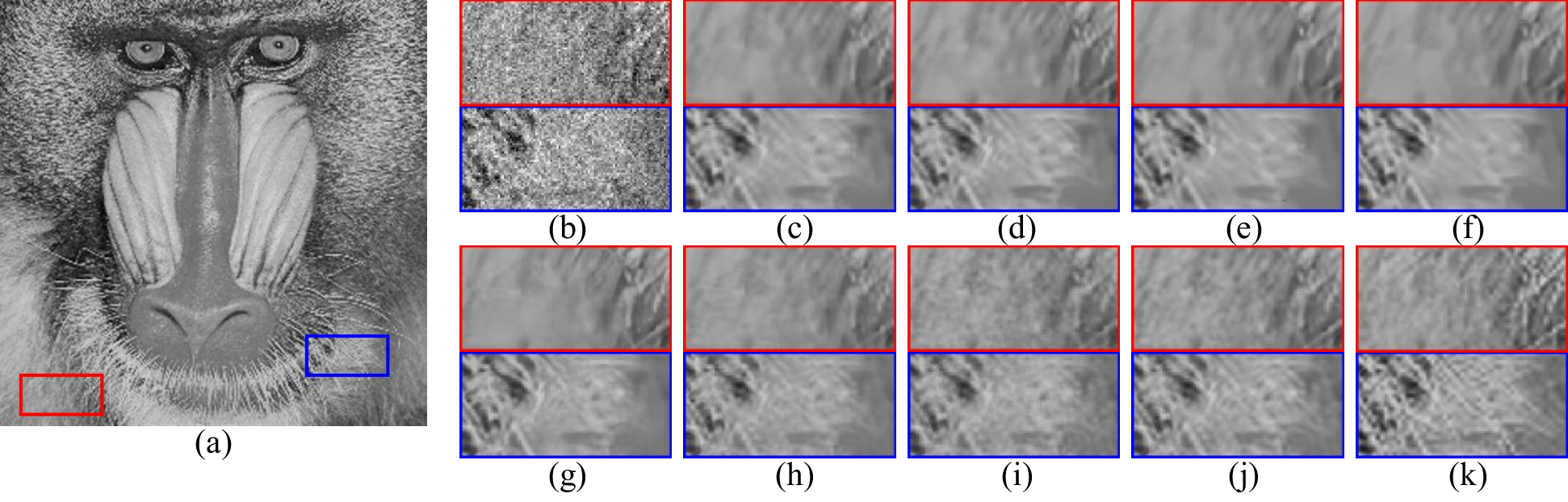}}
 \caption{\label{fig8} Denoising the Mandrill image at $\sigma =30$ with image details in the zoomed areas (red and blue boxes):(a)Mandrill image, (b)Noisy block, (c)BM3D, (d)BM3DSAPCA, (e)WNNM, (f)SLRD, (g)DnCNN-S, (h)SGHP, (i)AC-PT, (j)ACVA, (k)Noise-free block.}
\end{figure*}

\begin{figure*}[htb]
 \center{\includegraphics[width=18.0cm]  {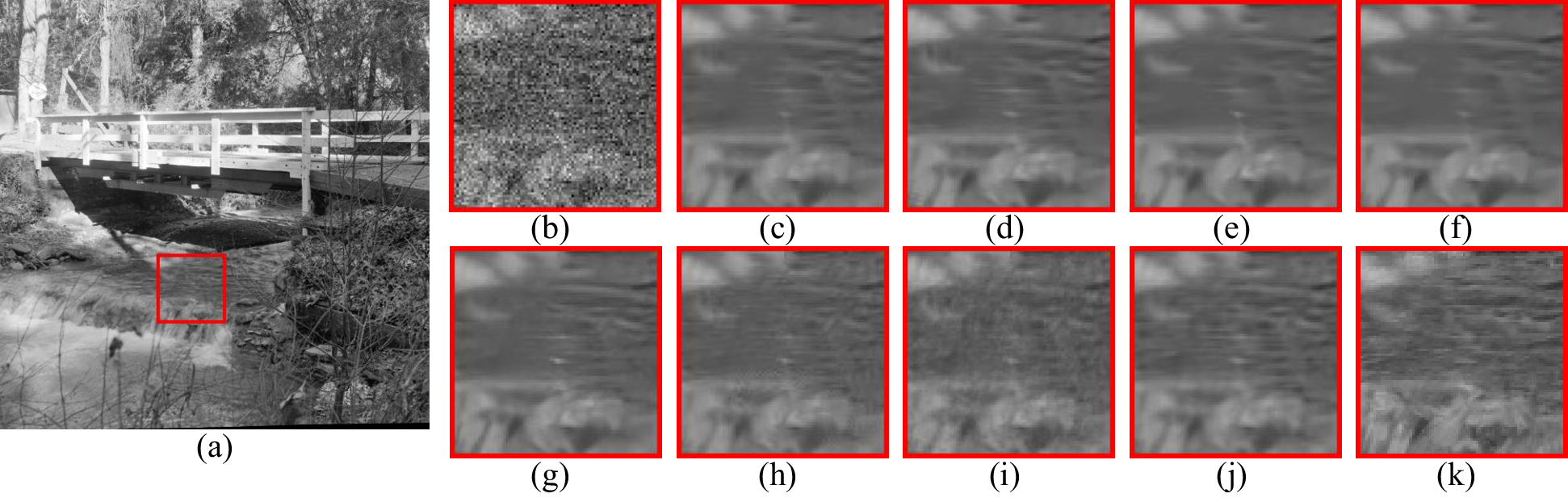}}
 \caption{\label{fig9} Denoising the Stream image at $\sigma =25$ with image details in the zoomed areas (red boxes):(a)Stream image, (b)Noisy block, (c)BM3D, (d)BM3DSAPCA, (e)WNNM, (f)SLRD, (g)DnCNN-S, (h)SGHP, (i)AC-PT, (j)ACVA, (k)Noise-free block.}
\end{figure*}

Fig. \ref{fig7}, \ref{fig8} and \ref{fig9} show that the proposed algorithm visually preserves textures details (such as the bark, baboon skin and water surface) better than other algorithms at different noise level.

\begin{table}[!t]
\scriptsize
\renewcommand{\arraystretch}{1.0}
\caption{ The average PSNR(dB), SSIM, FSIM results on 16 images with irregular textures from the USC-SIPI dataset. }
\label{tabe texture}
\centering
\begin{tabular}{|p{1.0cm}|p{0.5cm}|p{0.6cm}|p{0.6cm}|p{0.6cm}|p{0.6cm}|p{0.6cm}|}
\hline
Methods&$\sigma$&10&20&30&40&50\\\hline
\multirow{3}{*}{BM3D}&PSNR&31.55&27.83&25.91&24.44&23.59\\
&SSIM&0.9293&0.8430&0.7710&0.7044&0.6469\\
&FSIM&0.9882&0.9667&0.9432&0.9177&0.8925\\\hline
\multirow{3}{*}{SAPCA}&PSNR&31.62&27.88&25.92&24.62&23.65\\
&SSIM&0.9327&0.8500&0.7773&0.7141&0.6592\\
&FSIM&0.9884&0.9672&0.9436&0.9192&0.8961\\\hline
\multirow{3}{*}{SGHP}&PSNR&31.55&27.91&25.91&24.71&23.67\\
&SSIM&0.9314&0.8495&0.7752&0.7108&0.6443\\
&FSIM&0.9874&0.9653&0.9399&0.9147&0.8859\\\hline
\multirow{3}{*}{NCSR}&PSNR&31.58&27.79&25.80&24.50&23.54\\
&SSIM&0.9316&0.8453&0.7696&0.7004&0.6417\\
&FSIM&0.9879&0.9649&0.9381&0.9126&0.8842\\\hline
\multirow{3}{*}{WNNM}&PSNR&31.65&27.85&25.92&24.60&23.74\\
&SSIM&0.9311&0.8450&0.7726&0.7042&0.6557\\
&FSIM&0.9884&0.9659&0.9415&0.9121&0.8892\\\hline
\multirow{3}{*}{SLRD}&PSNR&31.64&27.90&25.98&24.70&23.78\\
&SSIM&0.9315&0.8504&0.7816&0.7164&0.6623\\
&FSIM&0.9883&0.9675&0.9438&0.9163&0.8894\\\hline
\multirow{3}{*}{DnCNN-S}&PSNR&\textbf{31.70}&27.94&25.97&24.66&23.73\\
&SSIM&0.9340&0.8545&0.7835&0.7199&0.6659\\
&FSIM&0.9884&0.9673&0.9430&0.9161&0.8903\\\hline
\multirow{3}{*}{AC-PT}&PSNR&31.65&27.89&25.96&24.63&23.59\\
&SSIM&0.9323&0.8499&0.7832&0.7257&0.6715\\
&FSIM&0.9884&0.9673&0.9455&0.9250&0.9047\\\hline
\multirow{3}{*}{ACVA}&PSNR&31.66&\textbf{27.98}&\textbf{26.07}&\textbf{24.81}&\textbf{23.87}\\
&SSIM&\textbf{0.9345}&\textbf{0.8554}&\textbf{0.7894}&\textbf{0.7331}&\textbf{0.6848}\\
&FSIM&\textbf{0.9886}&\textbf{0.9686}&\textbf{0.9482}&\textbf{0.9286}&\textbf{0.9103}\\\hline
\end{tabular}
\end{table}

\begin{center}
\begin{table}[h]
\scriptsize
\renewcommand{\arraystretch}{0.97}
\caption{PSNR(dB), SSIM and FSIM results of Gaussian denoising on six widely used test images. }
\label{tabe stdimage0}
\centerline{\resizebox{9.0cm}{!}{ %
\begin{threeparttable}[b]
\begin{tabular}{p{0.8cm}p{0.6cm}p{1.6cm}p{1.6cm}p{1.6cm}p{1.6cm}p{1.6cm}}%
\hline
Method                   & $\sigma$  & 10                                   & 20                                   &30                                   & 40                                   & 50             \\ \hline
\multirow{7}{*}{BM3D}&Fluocells&36.12$|$.9328$|$.9816&32.76$|$.8687$|$.9552&30.95$|$.8191$|$.9311&29.65$|$.7762$|$.9087&28.80$|$.7475$|$.8920\\
&House&36.71$|$.9218$|$.9549&33.77$|$.8726$|$.9225&32.09$|$.8480$|$.9065&30.65$|$.8256$|$.8917&29.69$|$.8122$|$.8763\\
&Lena&35.93$|$.9166$|$.9834&33.05$|$.8772$|$.9656&31.26$|$.8449$|$.9488&29.86$|$.8152$|$.9333&29.05$|$.7994$|$.9232\\
&Mandrill&33.14$|$.9327$|$.9813&29.07$|$.8503$|$.9508&26.85$|$.7741$|$.9214&25.27$|$.7028$|$.8929&24.38$|$.6433$|$.8682\\
&Stream&31.17$|$.9076$|$.9822&27.27$|$.7900$|$.9531&25.46$|$.6986$|$.9240&24.31$|$.6287$|$.8957&23.57$|$.5715$|$.8735\\
&Hill&33.67$|$.8851$|$.9784&30.76$|$.8061$|$.9522&29.18$|$.7525$|$.9304&28.03$|$.7098$|$.9101&27.22$|$.6770$|$.8941\\\hline
&Average&34.46$|$.9161$|$.9770&31.12$|$.8441$|$.9499&29.30$|$.7895$|$.9270&27.96$|$.7431$|$.9054&27.12$|$.7085$|$.8879\\\hline
\multirow{7}{*}{SAPCA}&Fluocells&36.16$|$.9334$|$.9818&32.78$|$.8665$|$.9548&30.95$|$.8133$|$.9295&29.68$|$.7686$|$.9066&28.71$|$.7329$|$.8867\\
&House&37.01$|$.9290$|$.9604&33.90$|$.8763$|$.9237&32.13$|$.8495$|$.9040&30.75$|$.8301$|$.8886&29.52$|$.8078$|$.8761\\
&Lena&36.07$|$.9183$|$.9838&33.20$|$.8809$|$.9668&31.40$|$.8505$|$.9518&30.10$|$.8247$|$.9378&29.08$|$.8014$|$.9236\\
&Mandrill&33.28$|$.9346$|$.9817&29.19$|$.8546$|$.9521&26.97$|$.7809$|$.9230&25.53$|$.7161$|$.8976&24.49$|$.6599$|$.8740\\
&Stream&31.36$|$.9119$|$.9829&27.51$|$.8028$|$.9550&25.64$|$.7117$|$.9267&24.50$|$.6410$|$.9018&23.70$|$.5861$|$.8786\\
&Hill&33.89$|$.8911$|$.9792&30.90$|$.8113$|$.9529&29.27$|$.7553$|$.9294&28.11$|$.7128$|$.9100&27.23$|$.6781$|$.8924\\\hline
&Average&34.63$|$.9197$|$.9783&31.25$|$.8487$|$.9509&29.39$|$.7935$|$.9274&28.11$|$.7489$|$.9071&27.12$|$.7110$|$.8886\\\hline
\multirow{7}{*}{SGHP}&Fluocells&35.79$|$.9209$|$.9799&32.68$|$.8701$|$.9540&30.69$|$.8176$|$.9287&29.54$|$.7823$|$.9051&28.54$|$.7479$|$.8836\\
&House&36.36$|$.9128$|$.9621&33.74$|$.8738$|$.9297&31.93$|$.8434$|$.9113&30.77$|$.8298$|$.8912&29.51$|$.8118$|$.8755\\
&Lena&35.65$|$.9089$|$.9823&32.88$|$.8737$|$.9641&30.94$|$.8379$|$.9463&29.82$|$.8211$|$.9341&28.70$|$.7970$|$.9206\\
&Mandrill&33.19$|$.9303$|$.9810&29.13$|$.8523$|$.9487&26.71$|$.7727$|$.9168&25.32$|$.7033$|$.8855&24.25$|$.6383$|$.8582\\
&Stream&31.22$|$.9084$|$.9818&27.38$|$.7998$|$.9513&25.50$|$.7128$|$.9220&24.36$|$.6386$|$.8944&23.56$|$.5772$|$.8704\\
&Hill&33.65$|$.8840$|$.9783&30.71$|$.8065$|$.9518&28.99$|$.7478$|$.9295&27.93$|$.7064$|$.9076&27.05$|$.6694$|$.8893\\\hline
&Average&34.31$|$.9109$|$.9775&31.09$|$.8460$|$.9499&29.13$|$.7887$|$.9258&27.96$|$.7469$|$.9030&26.94$|$.7069$|$.8829\\\hline
\multirow{7}{*}{NCSR}&Fluocells&36.07$|$.9329$|$.9811&32.70$|$.8713$|$.9539&30.80$|$.8234$|$.9284&29.54$|$.7824$|$.9029&28.58$|$.7516$|$.8809\\
&House&36.80$|$.9239$|$.9600&33.87$|$.8737$|$.9209&32.08$|$.8487$|$.8988&30.81$|$.8325$|$.8816&29.62$|$.8161$|$.8677\\
&Lena&35.85$|$.9157$|$.9821&32.95$|$.8768$|$.9629&31.06$|$.8455$|$.9442&29.92$|$.8239$|$.9318&28.90$|$.8035$|$.9182\\
&Mandrill&33.32$|$.9335$|$.9803&29.13$|$.8485$|$.9466&26.75$|$.7683$|$.9121&25.30$|$.6905$|$.8790&24.33$|$.6361$|$.8527\\
&Stream&31.20$|$.9059$|$.9813&27.32$|$.7890$|$.9496&25.52$|$.7024$|$.9190&24.32$|$.6198$|$.8915&23.54$|$.5677$|$.8669\\
&Hill&33.75$|$.8879$|$.9778&30.69$|$.8029$|$.9501&29.00$|$.7451$|$.9254&27.87$|$.6969$|$.9024&27.02$|$.6644$|$.8824\\\hline
&Average&34.50$|$.9166$|$.9771&31.11$|$.8437$|$.9473&29.20$|$.7889$|$.9213&27.96$|$.7410$|$.8982&27.00$|$.7066$|$.8781\\\hline
\multirow{7}{*}{WNNM}&Fluocells&36.17$|$.9337$|$.9820&32.74$|$.8687$|$.9537&30.93$|$.8213$|$.9271&29.61$|$.7806$|$.9014&28.75$|$.7539$|$.8818\\
&House&36.93$|$.9228$|$.9548&34.03$|$.8716$|$.9225&32.55$|$.8523$|$.9083&31.35$|$.8348$|$.8970&30.33$|$.8231$|$\textbf{.8836}\\
&Lena&36.05$|$.9177$|$.9827&33.12$|$.8787$|$.9634&31.43$|$.8502$|$.9468&30.11$|$.8220$|$.9303&29.25$|$.8059$|$.9183\\
&Mandrill&33.52$|$.9348$|$.9813&29.15$|$.8490$|$.9490&26.89$|$.7721$|$.9179&25.39$|$.6992$|$.8851&24.48$|$.6494$|$.8618\\
&Stream&31.26$|$.9080$|$.9826&27.38$|$.7925$|$.9529&25.59$|$.7053$|$.9240&24.46$|$.6315$|$.8939&23.69$|$.5815$|$.8719\\
&Hill&33.81$|$.8872$|$.9784&30.83$|$.8053$|$.9494&29.27$|$.7520$|$.9251&28.14$|$.7086$|$.9024&27.36$|$.6788$|$.8838\\\hline
&Average&34.62$|$.9174$|$.9770&31.21$|$.8443$|$.9485&29.44$|$.7922$|$.9249&28.17$|$.7461$|$.9017&27.31$|$.7154$|$.8835\\\hline
\multirow{7}{*}{SLRD}&Fluocells&36.12$|$.9324$|$.9815&32.84$|$.8720$|$.9542&31.01$|$.8273$|$.9268&29.80$|$.7910$|$.9002&28.87$|$\textbf{.7613}$|$.8787\\
&House&\textbf{37.06}$|$.9255$|$.9577&\textbf{34.14}$|$.8732$|$.9237&\textbf{32.68}$|$.8544$|$.9031&\textbf{31.58}$|$\textbf{.8435}$|$.8867&\textbf{30.64}$|$\textbf{.8333}$|$.8773\\
&Lena&36.10$|$.9192$|$.9831&33.27$|$.8818$|$.9648&31.51$|$\textbf{.8522}$|$.9461&\textbf{30.35}$|$\textbf{.8317}$|$.9297&\textbf{29.39}$|$\textbf{.8120}$|$.9167\\
&Mandrill&33.31$|$.9306$|$.9799&29.18$|$.8499$|$.9481&\textbf{27.01}$|$.7803$|$.9170&\textbf{25.58}$|$.7126$|$.8846&24.56$|$.6550$|$.8567\\
&Stream&31.25$|$.9061$|$.9818&27.44$|$.7957$|$.9526&25.62$|$.7097$|$.9236&24.48$|$.6367$|$.8945&23.68$|$.5801$|$.8663\\
&Hill&33.88$|$.8895$|$.9782&\textbf{30.92}$|$.8105$|$.9507&29.34$|$.7569$|$.9248&28.24$|$.7132$|$.8989&27.44$|$.6834$|$.8823\\\hline
&Average&34.62$|$.9172$|$.9770&31.30$|$.8472$|$.9490&29.53$|$.7968$|$.9236&\textbf{28.34}$|$.7548$|$.8991&\textbf{27.43}$|$.7208$|$.8797\\\hline
\multirow{7}{*}{DnCNN-S}&Fluocells&\textbf{36.28}$|$\textbf{.9363}$|$.9824&\textbf{32.95}$|$.8756$|$.9571&\textbf{31.10}$|$.8253$|$.9323&\textbf{29.89}$|$.7904$|$.9098&\textbf{28.87}$|$.7499$|$.8890\\
&House&36.52$|$.9115$|$.9556&33.89$|$.8702$|$.9225&32.30$|$.8509$|$.9061&31.02$|$.8341$|$.8907&30.02$|$.8200$|$.8794\\
&Lena&\textbf{36.20}$|$\textbf{.9195}$|$\textbf{.9842}&\textbf{33.40}$|$\textbf{.8836}$|$.9676&\textbf{31.59}$|$\textbf{.8546}$|$\textbf{.9525}&30.32$|$.8308$|$\textbf{.9389}&29.37$|$.8115$|$\textbf{.9277}\\
&Mandrill&\textbf{33.48}$|$\textbf{.9374}$|$.9823&\textbf{29.24}$|$\textbf{.8584}$|$.9533&26.98$|$\textbf{.7857}$|$.9243&25.55$|$.7213$|$.8961&24.57$|$.6685$|$.8714\\
&Stream&\textbf{31.49}$|$\textbf{.9159}$|$\textbf{.9831}&\textbf{27.73}$|$\textbf{.8166}$|$\textbf{.9563}&\textbf{25.91}$|$\textbf{.7354}$|$.9302&\textbf{24.77}$|$\textbf{.6673}$|$.9046&\textbf{23.97}$|$\textbf{.6133}$|$.8811\\
&Hill&\textbf{33.92}$|$.8923$|$.9797&30.91$|$.8103$|$.9537&\textbf{29.32}$|$.7556$|$.9309&\textbf{28.25}$|$.7149$|$.9110&\textbf{27.47}$|$\textbf{.6855}$|$.8947\\\hline
&Average&\textbf{34.65}$|$.9188$|$.9779&\textbf{31.35}$|$.8524$|$.9518&\textbf{29.54}$|$\textbf{.8013}$|$.9294&28.30$|$\textbf{.7598}$|$.9085&27.38$|$\textbf{.7248}$|$.8906\\\hline
\multirow{7}{*}{AC-PT}&Fluocells&36.03$|$.9307$|$.9816&32.62$|$.8658$|$.9566&30.58$|$.8001$|$.9286&29.29$|$.7633$|$.9103&27.92$|$.6852$|$.8777\\
&House&36.94$|$.9279$|$.9616&33.89$|$.8834$|$.9371&32.06$|$.8562$|$.9187&30.33$|$.8191$|$.9012&28.97$|$.7901$|$.8829\\
&Lena&35.76$|$.9124$|$.9830&32.58$|$.8630$|$.9630&30.62$|$.8218$|$.9423&29.13$|$.7787$|$.9208&27.86$|$.7360$|$.8936\\
&Mandrill&33.13$|$.9341$|$.9817&28.83$|$.8456$|$.9516&26.54$|$.7613$|$.9232&25.12$|$.6891$|$.9003&24.14$|$.6337$|$.8828\\
&Stream&30.93$|$.8989$|$.9814&27.17$|$.7856$|$.9525&25.37$|$.6977$|$.9262&24.32$|$.6376$|$.9048&23.53$|$.5899$|$.8852\\
&Hill&33.62$|$.8830$|$.9782&30.57$|$.8021$|$.9534&28.80$|$.7388$|$.9317&27.53$|$.6881$|$.9114&26.61$|$.6527$|$.8933\\\hline
&Average&34.40$|$.9145$|$.9779&30.94$|$.8409$|$.9523&29.00$|$.7793$|$.9284&27.62$|$.7293$|$.9081&26.51$|$.6813$|$.8859\\\hline
\multirow{7}{*}{ACVA}&Fluocells&36.21$|$.9353$|$\textbf{.9824}&32.88$|$\textbf{.8770}$|$\textbf{.9584}&31.03$|$\textbf{.8312}$|$\textbf{.9368}&29.74$|$\textbf{.7925}$|$\textbf{.9165}&28.76$|$.7603$|$\textbf{.8991}\\
&House&37.05$|$\textbf{.9309}$|$\textbf{.9631}&34.04$|$\textbf{.8859}$|$\textbf{.9359}&32.26$|$\textbf{.8580}$|$\textbf{.9132}&30.88$|$.8338$|$\textbf{.8981}&29.52$|$.8100$|$\textbf{.8835}\\
&Lena&36.07$|$.9190$|$.9841&33.15$|$.8808$|$\textbf{.9675}&31.32$|$.8490$|$.9517&29.94$|$.8208$|$.9365&28.89$|$.7960$|$.9223\\
&Mandrill&33.28$|$.9364$|$\textbf{.9824}&29.10$|$.8564$|$\textbf{.9543}&26.88$|$.7824$|$\textbf{.9278}&25.48$|$\textbf{.7220}$|$\textbf{.9064}&\textbf{24.64}$|$\textbf{.6815}$|$\textbf{.8926}\\
&Stream&31.26$|$.9119$|$.9826&27.53$|$.8063$|$.9556&25.69$|$.7202$|$\textbf{.9309}&24.56$|$.6585$|$\textbf{.9113}&23.76$|$.6084$|$\textbf{.8940}\\
&Hill&33.89$|$\textbf{.8925}$|$\textbf{.9796}&30.91$|$\textbf{.8150}$|$\textbf{.9558}&29.24$|$\textbf{.7596}$|$\textbf{.9362}&28.05$|$\textbf{.7152}$|$\textbf{.9183}&27.15$|$.6801$|$\textbf{.9027}\\\hline
&Average&34.63$|$\textbf{.9210}$|$\textbf{.9790}&31.27$|$\textbf{.8536}$|$\textbf{.9546}&29.40$|$.8001$|$\textbf{.9327}&28.11$|$.7571$|$\textbf{.9145}&27.12$|$.7227$|$\textbf{.8990}\\\hline
\end{tabular}
\end{threeparttable}}}
\end{table}
\end{center}

\begin{center}
\begin{table}[h]
\scriptsize
\renewcommand{\arraystretch}{0.97}
\caption{PSNR (dB), SSIM, FSIM results of Camera RAW image simulation. }
\label{tabe rawSim}
\centerline{\resizebox{9.0cm}{!}{ %
\begin{threeparttable}[b]
\begin{tabular}{p{0.8cm}p{0.6cm}p{1.6cm}p{1.6cm}p{1.6cm}p{1.6cm}p{1.6cm}}
\hline
Methods                    & $\alpha$ & 200   & 400    & 600   & 800  & 1000              \\\hline
\multirow{5}{*}{BM3D}&House&32.91$|$.8797$|$.9428&34.59$|$.9109$|$.9609&35.66$|$.9272$|$.9691&36.44$|$.9377$|$.9739&37.07$|$.9470$|$.9779\\
&Baboon&28.09$|$.8883$|$.9412&30.53$|$.9323$|$.9624&32.06$|$.9498$|$.9717&33.13$|$.9595$|$.9769&33.98$|$.9657$|$.9801\\
&Peppers&32.40$|$.8736$|$.9431&34.08$|$.9040$|$.9585&35.17$|$.9213$|$.9672&36.00$|$.9332$|$.9722&36.61$|$.9404$|$.9758\\
&Lena&32.67$|$.8810$|$.9456&34.39$|$.9068$|$.9591&35.40$|$.9209$|$.9659&36.16$|$.9303$|$.9705&36.75$|$.9373$|$.9739\\\hline
&Average&31.52$|$.8806$|$.9432&33.40$|$.9135$|$.9602&34.57$|$.9298$|$.9685&35.43$|$.9402$|$.9734&36.10$|$.9476$|$.9769\\\hline
\multirow{5}{*}{SAPCA}&House&\textbf{33.31}$|$.8914$|$.9503&\textbf{35.04}$|$\textbf{.9211}$|$\textbf{.9663}&\textbf{36.09}$|$\textbf{.9356}$|$\textbf{.9728}&\textbf{36.85}$|$\textbf{.9438}$|$\textbf{.9767}&\textbf{37.44}$|$\textbf{.9511}$|$.9799\\
&Baboon&\textbf{28.29}$|$.8966$|$.9440&\textbf{30.69}$|$.9361$|$.9635&\textbf{32.19}$|$.9520$|$.9725&\textbf{33.25}$|$.9611$|$.9775&\textbf{34.08}$|$\textbf{.9669}$|$.9806\\
&Peppers&\textbf{32.67}$|$.8784$|$.9465&\textbf{34.35}$|$.9091$|$.9612&\textbf{35.40}$|$.9254$|$.9694&\textbf{36.22}$|$.9365$|$.9738&\textbf{36.83}$|$.9432$|$.9771\\
&Lena&32.90$|$.8854$|$.9485&\textbf{34.62}$|$.9110$|$.9615&\textbf{35.63}$|$.9252$|$.9681&\textbf{36.37}$|$.9341$|$.9723&\textbf{36.96}$|$.9409$|$.9756\\\hline
&Average&\textbf{31.79}$|$.8879$|$.9473&\textbf{33.67}$|$.9193$|$.9631&\textbf{34.83}$|$.9345$|$.9707&\textbf{35.67}$|$.9439$|$.9751&\textbf{36.33}$|$.9505$|$.9783\\\hline
\multirow{5}{*}{NCSR}&House&33.10$|$.8883$|$.9483&34.77$|$.9175$|$.9643&35.80$|$.9323$|$.9709&36.56$|$.9407$|$.9748&37.18$|$.9491$|$.9785\\
&Baboon&28.10$|$.8825$|$.9382&30.53$|$.9285$|$.9607&32.06$|$.9469$|$.9704&33.14$|$.9570$|$.9760&33.98$|$.9634$|$.9794\\
&Peppers&32.42$|$.8747$|$.9446&34.12$|$.9048$|$.9587&35.18$|$.9207$|$.9668&36.02$|$.9327$|$.9718&36.63$|$.9392$|$.9750\\
&Lena&32.69$|$.8809$|$.9451&34.40$|$.9064$|$.9586&35.40$|$.9210$|$.9658&36.16$|$.9302$|$.9700&36.75$|$.9372$|$.9736\\\hline
&Average&31.58$|$.8816$|$.9440&33.46$|$.9143$|$.9606&34.61$|$.9302$|$.9685&35.47$|$.9402$|$.9732&36.13$|$.9472$|$.9766\\\hline
\multirow{5}{*}{SGHP}&House&32.68$|$.8807$|$.9486&34.26$|$.9112$|$.9635&35.21$|$.9259$|$.9696&35.89$|$.9359$|$.9734&36.45$|$.9441$|$.9771\\
&Baboon&28.12$|$.8909$|$.9445&30.51$|$.9328$|$.9634&32.01$|$.9499$|$.9720&33.05$|$.9595$|$.9771&33.84$|$.9654$|$.9802\\
&Peppers&32.38$|$.8765$|$.9484&34.09$|$.9071$|$.9610&35.10$|$.9228$|$.9681&35.89$|$.9338$|$.9726&36.44$|$.9401$|$.9756\\
&Lena&32.62$|$.8817$|$.9493&34.37$|$.9099$|$.9614&35.39$|$.9244$|$.9685&36.13$|$.9330$|$.9721&36.71$|$.9396$|$.9752\\\hline
&Average&31.45$|$.8825$|$.9477&33.31$|$.9152$|$.9623&34.43$|$.9307$|$.9696&35.24$|$.9405$|$.9738&35.86$|$.9473$|$.9770\\\hline
\multirow{5}{*}{SLRA}&House&33.25$|$.8815$|$.9459&34.91$|$.9130$|$.9634&35.98$|$.9301$|$.9716&36.74$|$.9398$|$.9758&37.37$|$.9485$|$.9797\\
&Baboon&28.10$|$.8803$|$.9353&30.58$|$.9294$|$.9600&32.08$|$.9470$|$.9695&33.14$|$.9570$|$.9752&33.98$|$.9636$|$.9787\\
&Peppers&32.63$|$.8755$|$.9431&34.30$|$.9063$|$.9592&35.33$|$.9219$|$.9670&36.14$|$.9332$|$.9718&36.72$|$.9395$|$.9751\\
&Lena&\textbf{32.92}$|$.8845$|$.9464&34.59$|$.9090$|$.9593&35.59$|$.9219$|$.9656&36.33$|$.9317$|$.9704&36.91$|$.9381$|$.9739\\\hline
&Average&31.72$|$.8804$|$.9427&33.60$|$.9144$|$.9605&34.75$|$.9302$|$.9684&35.59$|$.9404$|$.9733&36.24$|$.9474$|$.9768\\\hline
\multirow{5}{*}{WNNM}&House&33.24$|$.8835$|$.9443&34.93$|$.9143$|$.9636&35.96$|$.9305$|$.9709&36.71$|$.9392$|$.9752&37.33$|$.9481$|$.9790\\
&Baboon&28.20$|$.8888$|$.9409&30.64$|$.9331$|$.9622&32.15$|$.9501$|$.9715&33.22$|$.9597$|$.9768&34.06$|$.9659$|$.9801\\
&Peppers&32.58$|$.8753$|$.9438&34.22$|$.9047$|$.9594&35.27$|$.9213$|$.9677&36.09$|$.9332$|$.9727&36.69$|$.9400$|$.9761\\
&Lena&32.85$|$.8827$|$.9453&34.52$|$.9075$|$.9593&35.53$|$.9216$|$.9665&36.25$|$.9307$|$.9708&36.84$|$.9375$|$.9743\\\hline
&Average&31.72$|$.8826$|$.9435&33.58$|$.9149$|$.9611&34.73$|$.9309$|$.9692&35.57$|$.9407$|$.9739&36.23$|$.9479$|$.9774\\\hline
\multirow{5}{*}{DnCNN-S}&House&32.46$|$.8622$|$.9428&33.88$|$.8873$|$.9468&34.19$|$.8856$|$.9451&34.06$|$.8784$|$.9389&34.17$|$.8805$|$.9399\\
&Baboon&28.15$|$.8897$|$.9418&29.44$|$.8928$|$.9434&29.82$|$.8916$|$.9441&30.05$|$.8920$|$.9452&30.39$|$.8963$|$.9476\\
&Peppers&32.36$|$.8672$|$.9378&32.51$|$.8571$|$.9293&32.44$|$.8523$|$.9262&32.50$|$.8524$|$.9269&32.44$|$.8500$|$.9252\\
&Lena&32.79$|$.8788$|$.9423&33.92$|$.8967$|$.9544&34.30$|$.8969$|$.9537&34.53$|$.8975$|$.9536&34.50$|$.8942$|$.9515\\\hline
&Average&31.44$|$.8745$|$.9412&32.43$|$.8835$|$.9435&32.69$|$.8816$|$.9423&32.79$|$.8801$|$.9411&32.87$|$.8803$|$.9411\\\hline
\multirow{5}{*}{AC-PT}&House&32.95$|$.8905$|$.9530&34.68$|$.9181$|$.9647&35.78$|$.9331$|$.9724&36.62$|$.9428$|$.9761&37.17$|$.9492$|$.9787\\
&Baboon&28.14$|$.8929$|$.9438&30.60$|$.9342$|$.9634&32.06$|$.9506$|$.9721&33.14$|$.9602$|$.9773&34.03$|$.9663$|$\textbf{.9809}\\
&Peppers&32.43$|$.8796$|$\textbf{.9483}&34.16$|$.9090$|$.9616&35.26$|$.9249$|$.9689&36.06$|$.9358$|$.9737&36.69$|$.9427$|$.9767\\
&Lena&32.63$|$.8803$|$.9492&34.41$|$.9093$|$.9620&35.46$|$.9235$|$.9682&36.22$|$.9335$|$.9728&36.85$|$.9405$|$.9757\\\hline
&Average&31.54$|$.8858$|$.9486&33.46$|$.9177$|$.9629&34.64$|$.9330$|$.9704&35.51$|$.9431$|$.9750&36.18$|$.9497$|$.9780\\\hline
\multirow{5}{*}{ACVA}&House&33.01$|$\textbf{.8915}$|$\textbf{.9520}&34.73$|$.9206$|$.9661&35.75$|$.9349$|$.9726&36.54$|$.9435$|$.9765&37.13$|$.9507$|$\textbf{.9803}\\
&Baboon&28.20$|$\textbf{.8994}$|$\textbf{.9464}&30.57$|$\textbf{.9369}$|$\textbf{.9646}&32.06$|$\textbf{.9522}$|$\textbf{.9728}&33.12$|$\textbf{.9611}$|$\textbf{.9777}&33.96$|$.9669$|$.9807\\
&Peppers&32.45$|$\textbf{.8803}$|$.9481&34.16$|$\textbf{.9110}$|$\textbf{.9626}&35.23$|$\textbf{.9268}$|$\textbf{.9703}&36.06$|$\textbf{.9377}$|$\textbf{.9746}&36.66$|$\textbf{.9441}$|$\textbf{.9778}\\
&Lena&32.79$|$\textbf{.8860}$|$\textbf{.9504}&34.49$|$\textbf{.9118}$|$\textbf{.9630}&35.51$|$\textbf{.9262}$|$\textbf{.9696}&36.26$|$\textbf{.9349}$|$\textbf{.9735}&36.84$|$\textbf{.9414}$|$\textbf{.9765}\\\hline
&Average&31.61$|$\textbf{.8893}$|$\textbf{.9492}&33.49$|$\textbf{.9201}$|$\textbf{.9641}&34.64$|$\textbf{.9350}$|$\textbf{.9713}&35.49$|$\textbf{.9443}$|$\textbf{.9756}&36.15$|$\textbf{.9508}$|$\textbf{.9788}\\\hline
\end{tabular}
\end{threeparttable}}}
\end{table}
\end{center}

\subsection{Camera RAW image denoising}
\subsubsection{Camera RAW image simulation}
Four standard RGB test images, specifically, Peppers, Lena, Baboon, and House are selected for camera raw image denoising simulation. We adopt the simulation method used in \cite{yang2017evolutionary}.
To transform the RGB image into simulated raw images, we first scale the RGB image (within the bounds of $[0,1]$) to the domain of raw image:
\begin{equation}
\label{eqn_rgb2scale}
[r_{i,j}',g_{i,j}',b_{i,j}']^T=R_{max}\times [r_{i,j},g_{i,j},b_{i,j}]^T
\end{equation}
where $r$, $g$, and $b$ denote the signal in red, green, and blue channels, respectively. $i$ and $j$ are the x-coordinate and y-coordinate of any image pixel, respectively.

Then the pixels are further arranged into four subimages, $R$, $G1$, $G2$, and $B$ for simulation of the color filter array (CFA)\cite{Khashabi2014Joint}.
\begin{eqnarray}
\label{eqn_rgb2scale}
R&=&\{r_{i,j}''=r_{2i-1,2j-1}'\},\notag\\
G1&=&\{g_{i,j}''=g_{2i-1,2j}'\},\notag\\
G2&=&\{g_{i,j}''=g_{2i,2j-1}'\},\notag\\
B&=&\{b_{i,j}''=b_{2i,2j}'\}.
\end{eqnarray}
Finally, all the four subimages are arranged into a single image to simulate the RAW image.

To better simulate the noise in the real RAW image, we set the noise parameters $\alpha>0$,$b=0$,$p=0$ as \cite{yang2017evolutionary}. Based on Poisson-Gaussian noise reduction scheme described in Section 2, we apply the considered algorithm to denoising R, G1, G2, and B separately and then compute the average PSNR (SSIM and FSIM) results for evaluation.

\begin{figure*}[htb]
 \center{\includegraphics[width=18.1cm]  {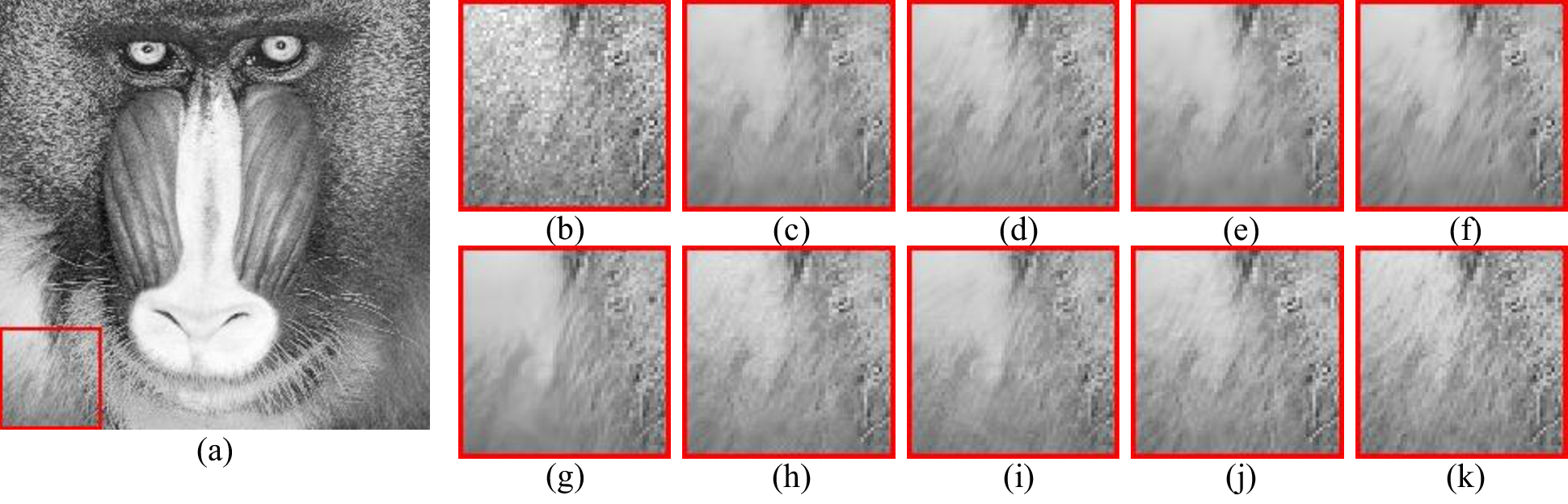}}
 \caption{\label{fig10} Performance comparison on the red channel of baboon image with image details in the zoomed areas (red boxes):(a)The red channel, (b)Noisy block ($\alpha=200$), (c)BM3D, (d)BM3DSAPCA, (e)NCSR, (f)WNNM, (g)DnCNN-S, (h)SGHP, (i)AC-PT, (j)ACVA, (k)Noise-free block.}
\end{figure*}

\begin{figure*}[htb]
 \center{\includegraphics[width=18.1cm]  {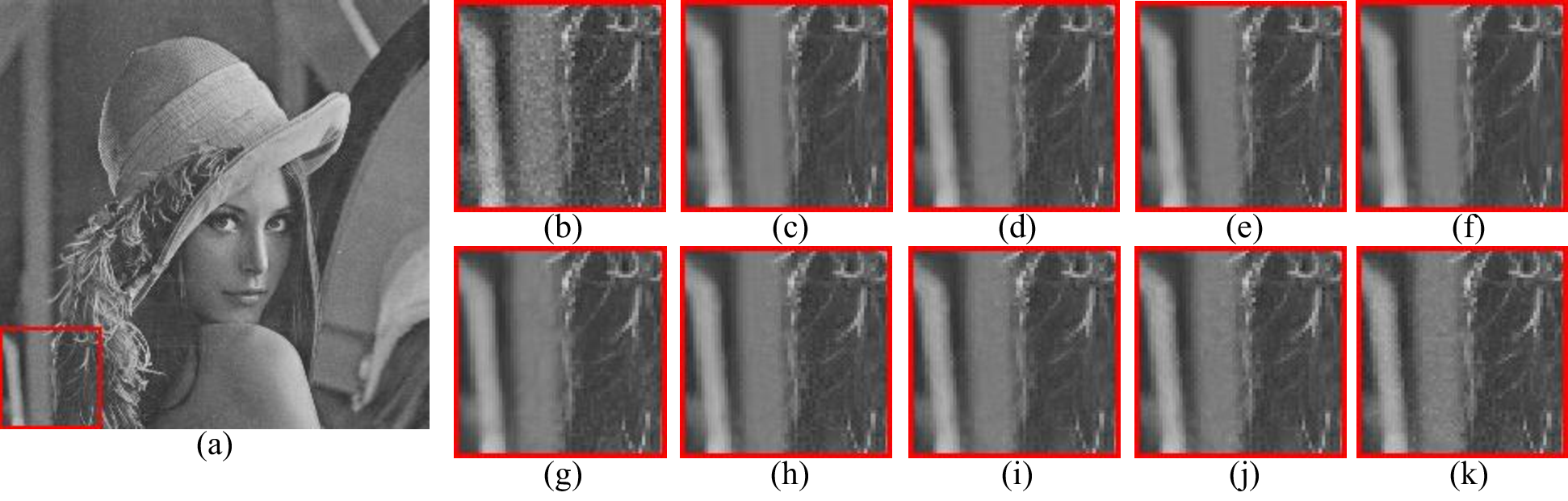}}
 \caption{\label{fig11} Performance comparison on the blue channel of Lena image with image details in the zoomed areas (red boxes):(a)The blue channel, (b)Noisy block ($\alpha=400$), (c)BM3D, (d)BM3DSAPCA, (e)NCSR, (f)WNNM, (g)DnCNN-S, (h)SGHP, (i)AC-PT, (j)ACVA, (k)Noise-free block.}
\end{figure*}

Table \ref{tabe rawSim} shows the superior SSIM and FSIM performance to all other considered algorithms, and better PSNR performance than that of BM3D, NCSR, SGHP and DnCNN-S. Since the denoising method EFBMD\cite{yang2017evolutionary} is inferior to NCSR in terms of the average PSNR results, we can conclude that ACVA is also better than EFBMD in terms of PSNR results. Fig. \ref{fig10} and \ref{fig11} show that ACVA preserves best the rough texture in the zoom-in area.

\subsubsection{Denoising on Real RAW Images}
\begin{figure*}[htb]
 \center{\includegraphics[width=18.1cm]  {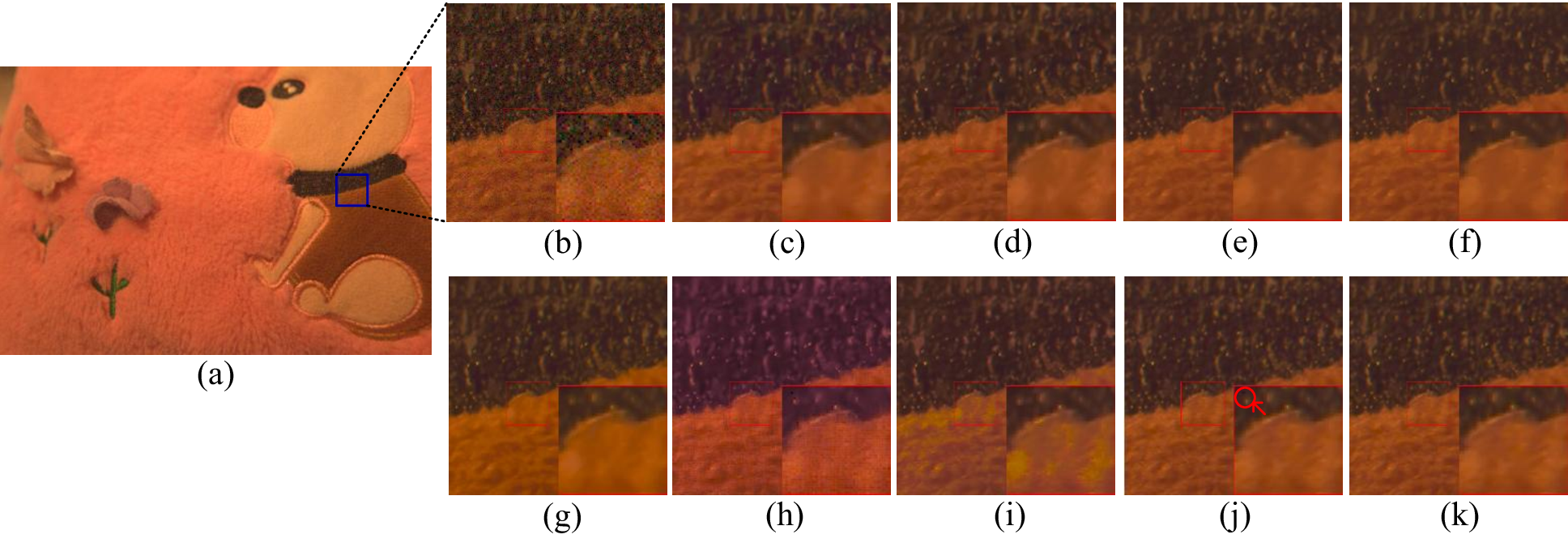}}
 \caption{\label{fig12} Denoising the real camera raw image with image details in the zoomed areas (red boxes): (a) Noisy image, (b) Small sample cut down from the blue square, (c) BM3D, (d) BM3DSAPCA, (e) NCSR, (f) WNNM, (g) SLRD , (h) DnCNN-S, (i) SGHP, (j) AC-PT, (k) ACVA.}
\end{figure*}

The RAW image of size $3744\times 5616$ is captured by a Canon EOS 5D Mark II. We cut down a $402\times 402$ square from the raw image for denoising tests. The noise parameters ($\alpha$ and $b$)in Poisson-Gaussian noise model are estimated by the method in \cite{foi2008practical}. We assume the noise level is invariant across the whole image. To avoid the over-estimate of noise level, we select the top-left $200\times 200$ flat area, estimate its R, G1, G2, B subimage separately, and adopt the minimum estimates of $\alpha$ and $b$, respectively.
After applying the GAT on the RAW image based on the estimated parameters, we denoise the real camera raw image using considered algorithms directly.
To visualize the denoised image, we adopt the method in \cite{Khashabi2014Joint} to transform the results into RGB images. Fig. \ref{fig12} shows that ACVA protects zoom-in details (such as singular points and textures) best compared with other algorithms. Specifically, we can also find that there is a noisy black dot mistakenly preserved by AC-PT. And serious color distortion can be observed in the results by SGHP and DnCNN, while BM3D, BM3DSAPCA, NCSR, SLRD, and WNNM just blur the isolated white points and brown texture. The serious color distortion by DnCNN implies that this state-or-the-art deep learning based denoising algorithm distort heavily the special textures resulted from the CFA, and how to control this kind of distortion remains an unsolved problem.

\section{Conclusions}
In this paper, we have proposed a texture preserving nonlocal denoising algorithm ACVA. In ACVA, an adaptive clustering method is designed to adaptively and robustly cluster similar patches. A state-of-the-art PCA-based denoising filter is proposed in a transform-domain texture variation adaptive filtering approach to perform a texture-preserving denoising of each cluster. The denoising performance of ACVA is further improved via a sliding window and aggregation approach. Experiments in Gaussian and camera raw image denoising demonstrate that the proposed denoising method ACVA achieves satisfactory texture-preserving denoising performance both quantitatively and visually.


%



\ifCLASSOPTIONcaptionsoff
  \newpage
\fi



\bibliographystyle{IEEEtran}
\bibliography{ref170327}

\begin{thebibliography}{10}
\providecommand{\url}[1]{#1}
\csname url@samestyle\endcsname
\providecommand{\newblock}{\relax}
\providecommand{\bibinfo}[2]{#2}
\providecommand{\BIBentrySTDinterwordspacing}{\spaceskip=0pt\relax}
\providecommand{\BIBentryALTinterwordstretchfactor}{4}
\providecommand{\BIBentryALTinterwordspacing}{\spaceskip=\fontdimen2\font plus
\BIBentryALTinterwordstretchfactor\fontdimen3\font minus
  \fontdimen4\font\relax}
\providecommand{\BIBforeignlanguage}[2]{{%
\expandafter\ifx\csname l@#1\endcsname\relax
\typeout{** WARNING: IEEEtran.bst: No hyphenation pattern has been}%
\typeout{** loaded for the language `#1'. Using the pattern for}%
\typeout{** the default language instead.}%
\else
\language=\csname l@#1\endcsname
\fi
#2}}
\providecommand{\BIBdecl}{\relax}
\BIBdecl

\bibitem{haidekker2011texture}
M.~A. Haidekker, ``Texture analysis,'' \emph{Advanced Biomedical Image
  Analysis}, pp. 236--275, 2011.

\bibitem{Texturesynthesis}
A.~A. Efros and T.~K. Leung, ``Texture synthesis by non-parametric sampling,''
  in \emph{Computer Vision, 1999. The Proceedings of the Seventh IEEE
  International Conference on}, vol.~2.\hskip 1em plus 0.5em minus 0.4em\relax
  IEEE, 1999, pp. 1033--1038.

\bibitem{BM3D}
K.~Dabov, A.~Foi, V.~Katkovnik, and K.~Egiazarian, ``Image denoising by sparse
  3-d transform-domain collaborative filtering,'' \emph{IEEE Transactions on
  Image Processing A Publication of the IEEE Signal Processing Society},
  vol.~16, no.~8, p. 2080, 2007.

\bibitem{LPG-PCA}
L.~Zhang, W.~Dong, D.~Zhang, and G.~Shi, ``Two-stage image denoising by
  principal component analysis with local pixel grouping,'' \emph{Pattern
  Recognition}, vol.~43, no.~4, pp. 1531--1549, 2010.

\bibitem{BM3DSAPCA}
K.~Dabov, A.~Foi, V.~Katkovnik, and K.~Egiazarian, ``Bm3d image denoising with
  shape-adaptive principal component analysis,'' in \emph{Proc Workshop on
  Signal Processing with Adaptive Sparse Structured Representation Saint-malo},
  2009.

\bibitem{WNNM}
S.~Gu, Q.~Xie, D.~Meng, W.~Zuo, X.~Feng, and L.~Zhang, ``Weighted nuclear norm
  minimization and its applications to low level vision,'' \emph{International
  Journal of Computer Vision}, vol. 121, no.~2, pp. 183--208, 2017.

\bibitem{SLRD}
M.~Nejati, S.~Samavi, H.~Derksen, and K.~Najarian, ``Denoising by low-rank and
  sparse representations,'' \emph{Journal of Visual Communication and Image
  Representation}, vol.~36, no.~C, pp. 28--39, 2016.

\bibitem{PLOW}
P.~Chatterjee and P.~Milanfar, ``Patch-based near-optimal image denoising,''
  \emph{IEEE Trans Image Process}, vol.~21, no.~4, pp. 1635--1649, 2012.

\bibitem{K-LLD}
------, ``Clustering-based denoising with locally learned dictionaries,''
  \emph{IEEE Transactions on Image Processing A Publication of the IEEE Signal
  Processing Society}, vol.~18, no.~7, p. 1438, 2009.

\bibitem{SGHP}
W.~Zuo, L.~Zhang, C.~Song, D.~Zhang, and H.~Gao, ``Gradient histogram
  estimation and preservation for texture enhanced image denoising.''
  \emph{IEEE Transactions on Image Processing A Publication of the IEEE Signal
  Processing Society}, vol.~23, no.~6, p. 2459, 2014.

\bibitem{acpt}
W.~Zhao, Y.~Lv, Q.~Liu, and B.~Qin, ``Detail-preserving image denoising via
  adaptive clustering and progressive pca thresholding,'' \emph{IEEE Access},
  vol.~6, no.~1, pp. 6303--6315, 2018.

\bibitem{LSSC}
J.~Mairal, F.~Bach, J.~Ponce, G.~Sapiro, and A.~Zisserman, ``Non-local sparse
  models for image restoration,'' in \emph{IEEE International Conference on
  Computer Vision}, 2010, pp. 2272--2279.

\bibitem{clusterPCA}
L.~Xu, J.~Li, Y.~Shu, and J.~Peng, ``Sar image denoising via clustering-based
  principal component analysis,'' \emph{IEEE Transactions on Geoscience and
  Remote Sensing}, vol.~52, no.~11, pp. 6858--6869, 2014.

\bibitem{Guo2016An}
Q.~Guo, C.~Zhang, Y.~Zhang, and H.~Liu, ``An efficient svd-based method for
  image denoising,'' \emph{IEEE Transactions on Circuits and Systems for Video
  Technology}, vol.~26, no.~5, pp. 868--880, 2016.

\bibitem{suboptimalwiener}
J.~Chen, J.~Benesty, Y.~Huang, and S.~Doclo, ``New insights into the noise
  reduction wiener filter,'' \emph{IEEE Transactions on Audio Speech and
  Language Processing}, vol.~14, no.~4, pp. 1218--1234, 2006.

\bibitem{foi2008practical}
A.~Foi, M.~Trimeche, V.~Katkovnik, and K.~Egiazarian, ``Practical
  poissonian-gaussian noise modeling and fitting for single-image raw-data,''
  \emph{IEEE Transactions on Image Processing}, vol.~17, no.~10, pp.
  1737--1754, 2008.

\bibitem{makitalo2013optimal}
M.~Makitalo and A.~Foi, ``Optimal inversion of the generalized anscombe
  transformation for poisson-gaussian noise,'' \emph{IEEE transactions on image
  processing}, vol.~22, no.~1, pp. 91--103, 2013.

\bibitem{GSURE}
J.~Bigot, C.~Deledalle, and D.~F{\'e}ral, ``Generalized sure for optimal shrinkage
  of singular values in low-rank matrix denoising,'' \emph{Journal of Machine
  Learning Research}, vol.~18, 2016.

\bibitem{bandwidth2014}
M.~Kohler, A.~Schindler, and S.~Sperlich, ``A review and comparison of
  bandwidth selection methods for kernel regression,'' \emph{International
  Statistical Review}, vol.~82, no.~2, pp. 243--274, 2014.

\bibitem{Qin2018}
B.~Qin, Z.~Shen, Z.~Fu, Z.~Zhou, Y.~Lv, and J.~Bao, ``Joint-saliency structure
  adaptive kernel regression with adaptive-scale kernels for deformable
  registration of challenging images,'' \emph{IEEE Access}, vol.~6, no.~1, pp.
  330--343, 2018.

\bibitem{LPAICI02}
V.~Katkovnik, ``A new method for varying adaptive bandwidth selection,''
  \emph{IEEE Transactions on Signal Processing}, vol.~47, no.~9, pp.
  2567--2571, 2002.

\bibitem{shapeadaptiveDCT}
A.~Foi, V.~Katkovnik, and K.~Egiazarian, ``Pointwise shape-adaptive dct for
  high-quality denoising and deblocking of grayscale and color images.''
  \emph{IEEE Transactions on Image Processing}, vol.~16, no.~5, p. 1395, 2007.

\bibitem{McGill}
A.~Olmos and F.~A. Kingdom, ``A biologically inspired algorithm for the
  recovery of shading and reflectance images.'' \emph{Perception}, vol.~33,
  no.~12, p. 1463, 2004.

\bibitem{PSNR}
Q.~Huynh-Thu and M.~Ghanbari, ``Scope of validity of psnr in image/video
  quality assessment,'' \emph{Electronics Letters}, vol.~44, no.~13, pp.
  800--801, 2008.

\bibitem{SSIM}
Z.~Wang, A.~C. Bovik, H.~R. Sheikh, and E.~P. Simoncelli, ``Image quality
  assessment: from error visibility to structural similarity,'' \emph{IEEE
  Transactions on Image Processing A Publication of the IEEE Signal Processing
  Society}, vol.~13, no.~4, pp. 600--612, 2004.

\bibitem{FSIM}
L.~Zhang, L.~Zhang, X.~Mou, and D.~Zhang, ``Fsim: A feature similarity index
  for image quality assessment,'' \emph{IEEE Transactions on Image Processing A
  Publication of the IEEE Signal Processing Society}, vol.~20, no.~8, p. 2378,
  2011.

\bibitem{Zhang2017Beyond}
K.~Zhang, W.~Zuo, Y.~Chen, D.~Meng, and L.~Zhang, ``Beyond a gaussian denoiser:
  Residual learning of deep cnn for image denoising.'' \emph{IEEE Transactions
  on Image Processing}, vol.~PP, no.~99, pp. 1--1, 2017.

\bibitem{hardthrehold-svd}
M.~Gavish and D.~L. Donoho, ``The optimal hard threshold for singular values is
  $4/\sqrt {3}$,'' \emph{Information Theory IEEE Transactions on}, vol.~60,
  no.~8, pp. 5040--5053, 2014.

\bibitem{optimalshrinkSVD16}
M.~Gavish and D.~Donoho, ``Optimal shrinkage of singular values,'' \emph{IEEE
  Transactions on Information Theory}, vol.~PP, no.~99, pp. 1--1, 2014.

\bibitem{yang2017evolutionary}
C.-C. Yang, S.-M. Guo, and J.~S.-H. Tsai, ``Evolutionary fuzzy
  block-matching-based camera raw image denoising,'' \emph{IEEE transactions on
  cybernetics}, vol.~47, no.~9, pp. 2862--2871, 2017.

\bibitem{Khashabi2014Joint}
D.~Khashabi, S.~Nowozin, J.~Jancsary, and A.~W. Fitzgibbon, ``Joint demosaicing
  and denoising via learned nonparametric random fields.'' \emph{Image
  Processing IEEE Transactions on}, vol.~23, no.~12, pp. 4968--81, 2014.

\end{thebibliography}
\end{document}